\pdfoutput=1

\documentclass[11pt]{article}

\usepackage[preprint]{acl}

\usepackage{times}
\usepackage{latexsym}

\usepackage[T1]{fontenc}

\usepackage[utf8]{inputenc}

\usepackage{microtype}

\usepackage{inconsolata}

\usepackage{graphicx}

\usepackage{colortbl}
\usepackage{multirow, booktabs}
\usepackage{multicol}
\usepackage{amsmath}
\usepackage{hyperref}
\usepackage{upgreek}

\usepackage{tcolorbox}
\usepackage{lipsum} 
\newcommand{\VarSty}[1]{\textnormal{\ttfamily\color{blue!90!black}#1}\unskip}
\title{\textsc{Chain-of-Talkers (CoTalk)}: \\ Fast Human Annotation of Dense Image Captions}

\author{
\textbf{Yijun Shen\textsuperscript{1}}$^*$\quad
\textbf{Delong Chen\textsuperscript{2}}$^*$\quad
\textbf{Fan Liu\textsuperscript{1,†}}\quad
\textbf{Xingyu Wang\textsuperscript{1}}\quad\\
\textbf{Chuanyi Zhang\textsuperscript{1}}\quad
\textbf{Liang Yao\textsuperscript{1}}\quad
\textbf{Yuhui Zheng\textsuperscript{1,†}}\\
\textsuperscript{1}Hohai University \textsuperscript{2}HKUST\\
\texttt{fanliu@hhu.edu.cn},\texttt{zhengyh@vip.126.com}
}

\begin{document}
\maketitle
\begin{abstract}
While densely annotated image captions significantly facilitate the learning of robust vision-language alignment, methodologies for systematically optimizing human annotation efforts remain underexplored. We introduce \textsc{\textbf{Chain-of-Talkers (CoTalk)}}, an AI-in-the-loop methodology designed to maximize the number of annotated samples and improve their comprehensiveness under fixed budget constraints (\textit{e.g.,} total human annotation time). The framework is built upon two key insights. First, sequential annotation reduces redundant workload compared to conventional parallel annotation, as subsequent annotators only need to annotate the \textit{``residual''}---the missing visual information that previous annotations have not covered. Second, humans process textual input faster by reading while outputting annotations with much higher throughput via talking; thus a multimodal interface enables optimized efficiency. We evaluate our framework from two aspects: intrinsic evaluations that assess the comprehensiveness of \textit{semantic units}, obtained by parsing detailed captions into object-attribute trees and analyzing their effective connections; extrinsic evaluation measures the practical usage of the annotated captions in facilitating vision-language alignment. Experiments with eight participants show our \textsc{Chain-of-Talkers} (CoTalk) improves annotation speed (0.42 vs. 0.30 units/sec) and retrieval performance (41.13\% vs. 40.52\%) over the parallel method.
\end{abstract}

\section{Introduction}

\begin{figure}
\centering
\includegraphics[width=1\linewidth]{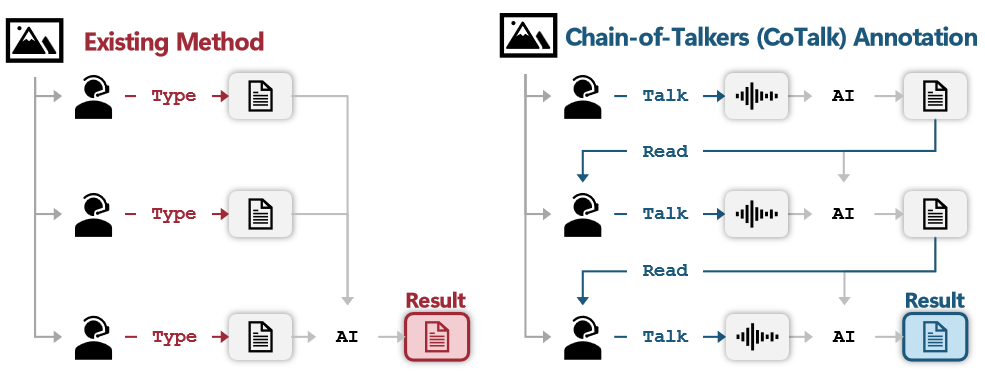}
\caption{Comparison between Existing Annotation Frameworks and \textsc{\textbf{Chain-of-Talkers (CoTalk)}}: Traditional methods require annotators to independently type complete image descriptions. In contrast, \textsc{CoTalk} has the first annotator provide a complete spoken annotation, while subsequent annotators focus solely on the \textit{``residual''}—the overlooked details.}
\label{fig:The comparsion between existing annotation framework and Chain-of-Talkers framework}
\end{figure}

Language is central to human–AI interaction, while vision enables high-bandwidth perception of the world. Aligning these modalities semantically is essential for AI systems to interpret multimodal information effectively. Typically, this alignment is achieved by learning from images paired with human-generated captions. Early datasets, such as COCO~\cite{Chen2015MicrosoftCC}, provide brief single-sentence captions per image, offering basic alignment but limited comprehensiveness. Recently, research has shifted toward creating denser captions, significantly improving vision-language models through enhanced semantic grounding, interpretability, and downstream task performance~\cite{Cho2025PerceptionLMOD, Shabbir2025GeoPixelPG}.

Currently, annotation interfaces commonly display images to annotators who examine visual content and generate captions by typing~\cite{Garg2024ImageInWordsUH, Hua2024FINECAPTIONCI}. Parallel annotation, wherein multiple annotators independently describe the same image, aims to boost comprehensiveness by aggregating diverse perspectives~\cite{Deitke2024MolmoAP, Athar2024ViCaSAD, Onoe2024DOCCIDO, Hu2024CanLD}. Despite their simplicity and ease of use, these methods are largely heuristic, lacking rigorous theoretical justification or systematic validation. In this paper, we identify two fundamental limitations of current annotation practices and propose a novel, systematically validated methodology to overcome these issues.

Our first insight addresses the inefficiency caused by redundancy in parallel annotations, where multiple annotators independently describe the same visual content, leading to substantial overlap (\textit{e.g.,} PixomoCap~\cite{Deitke2024MolmoAP} cross-annotation overlap is 71.36\% as measured by Sentence-BERT~\cite{Reimers2019SentenceBERTSE}). Our second insight is that speech-based annotation significantly surpasses typing in speed and efficiency. This is supported by prior research demonstrating that the average throughput for spoken words (161.2 words per minute, WPM) substantially exceeds that of typing (53.46 WPM)~\cite{2016Speech}.

Building on these insights, we introduce \textsc{\textbf{Chain-of-Talkers (CoTalk)}}, an novel AI-in-the-loop annotation framework (illustrated in Figure~\ref{fig:The comparsion between existing annotation framework and Chain-of-Talkers framework}). Annotators sequentially contribute descriptions: the first provides a comprehensive initial description from scratch, while subsequent annotators incrementally add only the \textit{``residual''}, the missing visual details. Each annotator reads previous annotations and communicates additional details through speech, which are automatically transcribed and synthesized into coherent text by a Large Language Model (LLM). This sequential, multimodal process significantly reduces redundancy and maximizes annotation comprehensiveness.

The design of \textsc{CoTalk} is theoretically grounded in an information-theoretic evaluation framework for image captions~\cite{Chen2024WhatMF}, which provides rigorous guidelines for developing efficient annotation methodologies. To validate our method, we conduct comprehensive assessments of human-generated annotations through both intrinsic and extrinsic evaluations.

\textbf{Intrinsic Evaluation}. Traditional metrics such as caption length are biased by linguistic style, and model-based measures (\textit{e.g.,} CLIP-Score~\cite{Hessel2021CLIPScoreAR}, CLIP-IMAGE-Score~\cite{Ge2024VisualFC}) strongly depend on the performance of the underlying model, limiting their interpretability. To overcome this, we introduce an alternative evaluation based on semantic units—object-attribute pairs extracted by LLMs from detailed captions to construct semantic trees. The number of effective object-attribute connections represents annotation comprehensiveness. Using this metric, we demonstrate that \textsc{CoTalk} achieves superior annotation comprehensiveness, generating 36.72 semantic units per image compared to 33.61 with parallel annotation. It also reduces annotation time by approximately 48\%, increasing the annotation speed from 0.30 to 0.42 semantic units per second. Additionally, we verify that speech-based annotation is not only faster but also captures richer detail compared to typing, and reading previous annotations leads to better comprehension than auditory reviews alone. Since speech-based annotation reaches a speed of 0.40 semantic units per second, significantly faster than typing at 0.17. Meanwhile, reading prior annotations takes 55.20 seconds with 100\% accuracy, compared to 70.20 seconds and 94\% accuracy for auditory reviews.

\textbf{Extrinsic Evaluation}. To validate the practical utility of \textsc{CoTalk}, we employ retrieval-based evaluation, a robust proxy for annotation quality relevant to real-world vision-language tasks. By fine-tuning a CLIP model~\cite{Radford2021LearningTV} separately using captions generated via \textsc{CoTalk} and parallel annotation methods, we systematically compare their downstream retrieval performances across multiple benchmarks. Our results demonstrate that \textsc{CoTalk} consistently yields superior retrieval accuracy, achieving an average of 41.13\% across three datasets and six retrieval tasks, surpassing parallel annotations (40.52\%). This highlights \textsc{CoTalk}'s ability to produce more semantically rich and practically valuable annotations, underscoring its effectiveness in enhancing vision-language model performance.

\section{Methodology}

\begin{figure*}
    \centering
    \includegraphics[height=5cm]{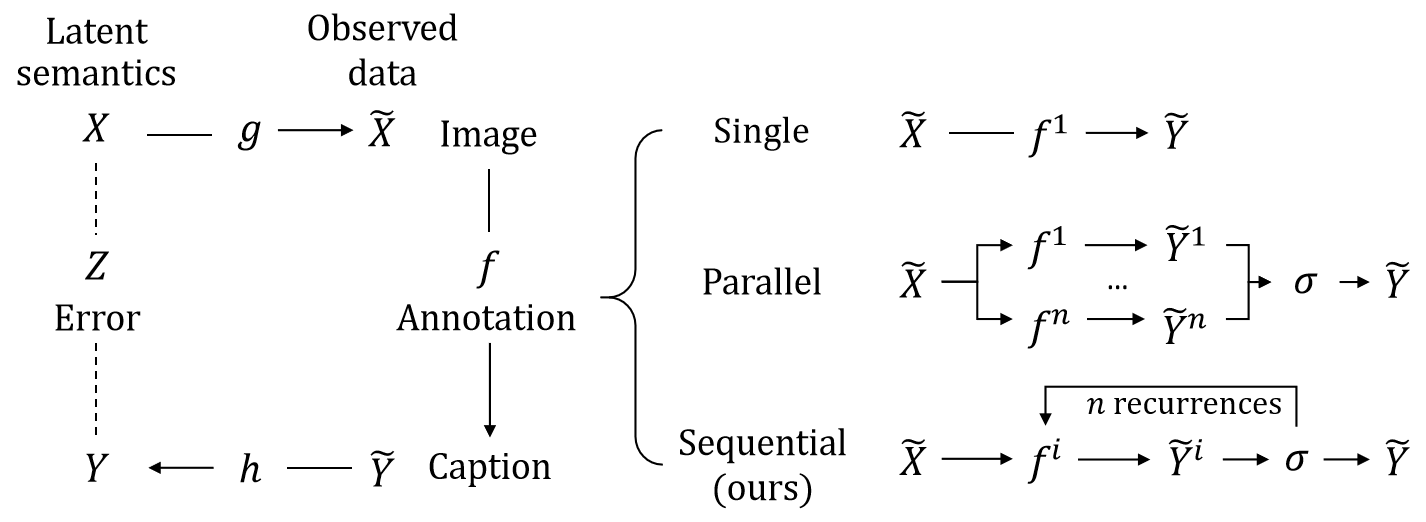}
    \caption{Overview of our formulation. Some latent variable ${X}$ in a latent semantic space ${S}$ generates image $\widetilde{{X}}$ in data space ${D}$. The image ${X}$ is then annotated by ${f}$ producing a caption $\widetilde{{Y}}$ which can be mapped back to the original latent space as ${Y}$. The semantic-level error measures the annotation quality. The annotation function ${f}$ can take several forms: in single annotation, a single annotator provides a complete image description; in parallel annotation, multiple annotators independently generate captions that are later merged; and in our sequential annotation, the first annotator provides an initial description, with subsequent annotators incrementally enriching it.}
    \label{fig:method}
\end{figure*}

\subsection{Preliminary}
\label{Preliminary}
 ~\citet{Chen2024WhatMF} propose an information-theoretic framework for image captioning, systematically defining key criteria for high-quality annotations (Figure \ref{fig:method}). The framework introduces a semantic space ${S} \in [{0,1}]^n$, where each dimension corresponds to a semantic unit $\omega_{{i}}$ in {$\Omega = \left\{\omega_{i}\right\}_{i-1}^n$}, representing the probability of that unit appearing in the image. The caption generation process is modeled as a function ${f}^\theta$, simulating human annotation. For ${n}$ annotators, individual annotation processes are denoted ${f}^1, \dots, {f}^n$, producing annotations $\widetilde{{Y}}^k$ for the ${k}$-th annotator. These annotations are mapped to the semantic space as vectors ${Y}$ via a transformation ${h}(\cdot)$. In this binary semantic space, a value of 1 indicates the presence of a semantic unit, while 0 indicates its absence. The overlap between the source semantics ${X}$ and received semantics ${Y}$ defines their semantic consistency. The semantic-level error is given by ${Z = Y - X}$, capturing the discrepancy between the intended semantics and the annotated interpretation, thereby reflecting annotation quality.

~\citet{Chen2024WhatMF} then proposes three core objectives to guide high-quality captioning: 

(1) \textbf{Information Sufficiency}. ${J}_{{\text{suf}}}(\theta) = {I(Y; X)}$: ensures the caption captures task-relevant semantics by maximizing mutual information;

(2) \textbf{Minimal Redundancy}. ${J}_{\text{min}}(\theta) = {-H}({\widetilde{{Y}}})$: promotes conciseness by minimizing entropy, reducing repetitive or irrelevant content;

(3) \textbf{Human Comprehensibility}. ${J}_{\text{int}}(\theta) = {-D(P}_{{\widetilde{{Y}}}}\left|\right|{P}_{\text{lang}})$: measures alignment with natural language through distribution distance, enhancing readability. 
These objectives are integrated into a unified optimization function:
\begin{equation}
	{J}(\theta) = {J}_{\text{suf}}(\theta) - \beta {J}_{\text{min}}(\theta) - \gamma {J}_{\text{int}}(\theta),\label{equation0}
\end{equation}
where $\beta$ and $\gamma$ are tunable weights. This formulation offers a flexible and principled approach to both analyzing and optimizing image captioning systems.

\subsection{Human Annotation}
Traditional human annotation typically involves a single annotator, denoted as ${f}^{{1}}$, producing the annotation $\widetilde{{Y}}^{{1}}$. Thus, the result of single-round annotation is defined as:
\begin{equation}
	\widetilde{{Y}}_{\text{single}} = \widetilde{{Y}}^{1}.\label{equation_Y_single}
\end{equation}
In addition to annotation content, time is a crucial factor. The total annotation time in a single round comprises two components: ${T}_{\text{in}}^{\widetilde{{X}}}$, the time spent observing the image, and ${T}_{\text{out}}$, the time spent generating the annotation. The total time can be expressed as: ${{T}}_{\text{single}} = {T}_{\text{in}}^{\widetilde{{X}}} + {T}_{\text{out}}^{\widetilde{{Y}}^{{1}}}$.

\subsection{Parallel Annotation}

Due to the high reliance on a single annotator, single-round annotation often results in inconsistent quality. To address this, \textit{parallel annotation} is introduced, where multiple annotators work independently to produce separate outputs~\cite{Deitke2024MolmoAP,Onoe2024DOCCIDO}. These outputs are then aggregated into a unified annotation using a merging function. We define this aggregation process as the \textbf{LLM merger} $\sigma(\cdot)$, which consolidates individual annotations into a single, coherent result.

Formally, ${f}^{{k}}$ (${k = 1}, \dots{, n}$) generates a description based directly on the image. The annotation is denoted as $\widetilde{{Y}}^{{k}}$.

After all n annotators have completed their annotations, the LLM merges all the annotations from the different annotators together.
\begin{equation}
	\widetilde{{Y}}_{\text{Par}} = \widetilde{{Y}}_{\sigma}^{{n}} = \sigma (\widetilde{{Y}}^{{1}}, \dots, \widetilde{{Y}}^{{n}}),\label{equation_p}
\end{equation}
then the final semantic unit of the parallel process can be determined as ${Y}_{\text{Par}} = {Y}_{\sigma}^{{n}}={h}(\widetilde{{Y}}_{\sigma}^{{n}})$

The total time cost for the parallel annotation process is: ${T}_{\text{Par}} = {n} \cdot {T}_{\text{in}}^{\widetilde{{X}}} + \sum_{{i=1}}^{{n}}{T}_{\text{out}}^{\widetilde{{Y}}^{{i}}}$.

\subsection{Chain-of-Talkers Annotation}
We introduce the Chain-of-Talkers (CoTalk) method, designed to reduce human annotation time while maintaining high annotation quality. CoTalk is according to two key insights: (1) sequential annotation can be more efficient than parallel annotation, and (2) generating annotations via speech (talk) is faster than typing, while comprehending prior annotations is quicker through text than audio.

At a system level, CoTalk adopts a sequential annotation strategy, where only the first annotator describes the entire image, and subsequent annotators contribute only residual information—i.e., additions or corrections according to what has already been annotated. At the individual level, CoTalk leverages a cross-modal approach: prior annotations are consumed as text, and new annotations are generated through talk. This framework is illustrated in Figure~\ref{fig:The comparsion between existing annotation framework and Chain-of-Talkers framework}.

Formally, ${f}^{{k}}$ (${k = 2,} \dots{, n}$) generates a talk description based on previous annotation produced by ${f}^{{k-1}}$ (when ${k = 1}$, the description is based directly on the image). After converting the talk to text, the resulting annotation is denoted as $\widetilde{{Y}}^{{k}}$. 
\begin{equation}
	\widetilde{{Y}}^{{k}} = {f}^{{k}}(\widetilde{{Y}}_{\sigma}^{{k-1}}, \widetilde{{X}}),\label{equation1}
\end{equation}

After each annotator completes their annotation, the LLM merges the current annotation with the previously aggregated annotations.

\begin{equation}
	\widetilde{{Y}}_{\text{CoTalk}} = \widetilde{{Y}}_{\sigma}^{{k}} = \sigma (\widetilde{{Y}}_{\sigma}^{{k-1}}, \widetilde{{Y}}^{{k}}),\label{equation2}
\end{equation}

This process continues until an annotator determines that no further information is necessary and the annotation is complete.

To account for time, we introduce ${T}_{\text{in}}^{\widetilde{{Y}}_{\sigma}^{{k-1}}}$,
representing the time of ${f}^{{k}}$ required to read the previous annotation. The total time cost for the CoTalk method is: ${T}_{\text{CoTalk}} = {n} \cdot {T}_{\text{in}}^{\widetilde{{X}}} + \sum_{{i=1}}^{{n}}({T}_{\text{in}}^{\widetilde{{Y}}_{\sigma}^{{i-1}}} + {T}_{\text{out}}^{\widetilde{{Y}}^{{i}}})$.

\section{Theoretical Analysis}

\subsection{CoTalk is Pareto Optimal}
In this section, we compare the CoTalk method with parallel annotation and one-round human annotation to highlight its advantages in both quality and efficiency. Annotation quality is evaluated using ${J}_{\theta}$(defined in Equation~\ref{equation0}), while efficiency is quantified as:
\begin{equation}
{E} = \frac{{J}(\theta)}{{T}},\label{equation_Efficiency}
\end{equation}
where ${T}$ denotes the total annotation time. Then, we first demonstrate that CoTalk achieves higher informational sufficiency with minimal redundancy, indicating superior annotation quality. Subsequently, we show that for annotations of comparable quality, CoTalk requires a lower annotation budget, thereby achieving Pareto optimality in the quality-efficiency trade-off.

Before proceeding with the analysis, we introduce the following assumption:

\textbf{Assumption 1 (Diminishing semantic contribution in CoTalk annotations)} 
In CoTalk, the amount of new semantic content added by each successive annotator decreases due to the influence of prior annotations, resulting in ${Y}_{\text{CoTalk}}^{{k}} > {Y}_{\text{CoTalk}}^{{k+1}}$.

\textbf{Assumption 2 (Effective and Correlated Merging of Annotations)} 
In both CoTalk and parallel annotation, the merging function $\sigma(\cdot)$ effectively eliminates redundancy and accurately integrates semantic units. In addition, it exhibits a positive input-output correlation: greater input yields more output, as shown in~\ref{app:support1}.

We now present theorems according to the assumptions: 

\textbf{Theorem 1 (CoTalk Enhances Annotation Quality)}: As defined in Equation~\ref{equation0}, high quality annotation is characterized by high information sufficiency, minimal redundancy, and strong human comprehensibility.

\textbf{Information Sufficiency}: Information sufficiency is defined as the completeness of semantic unit coverage, representing how thoroughly annotations capture the semantic content of an image. Under Assumptions 1-2, CoTalk demonstrates superior information sufficiency compared to single-round and parallel annotation. In single-round annotation, a single annotator provides limited semantic coverage. While parallel annotation improves coverage by aggregating multiple independent annotations, it suffers from redundancy, as each annotator samples from the entire semantic space. In contrast, CoTalk allows each annotator to build on the previous ones; the ${f}^{{k}}$ samples from the residual semantic space ${Y -} {{Y}_\sigma^{{k-1}}}$, focusing only on what has not yet been covered. This targeted supplementation reduces redundancy and increases semantic completeness. Consequently, the incremental information gain per round, defined as {$\Delta {I}({{Y}}_{\sigma}^{{k}};{X}) = {I}({{Y}}_{\sigma}^{{k}}{;X}) - {I}({{Y}}_{\sigma}^{{k-1}}{;X)}$}, is greater for CoTalk than for parallel annotation when ${k>=2}$:

\begin{equation}
	\Delta {I}_{\text{CoTalk}}({{Y}}_\sigma^{{k}}{;X}) > \Delta {I}_{\text{par}}({{Y}}_{\sigma}^{k};{X}) \label{equation_info}
\end{equation}
This indicates that CoTalk accumulates semantic information more efficiently over successive rounds, thereby achieving higher information sufficiency.

\begin{figure}
    \centering
    \includegraphics[width=1\linewidth]{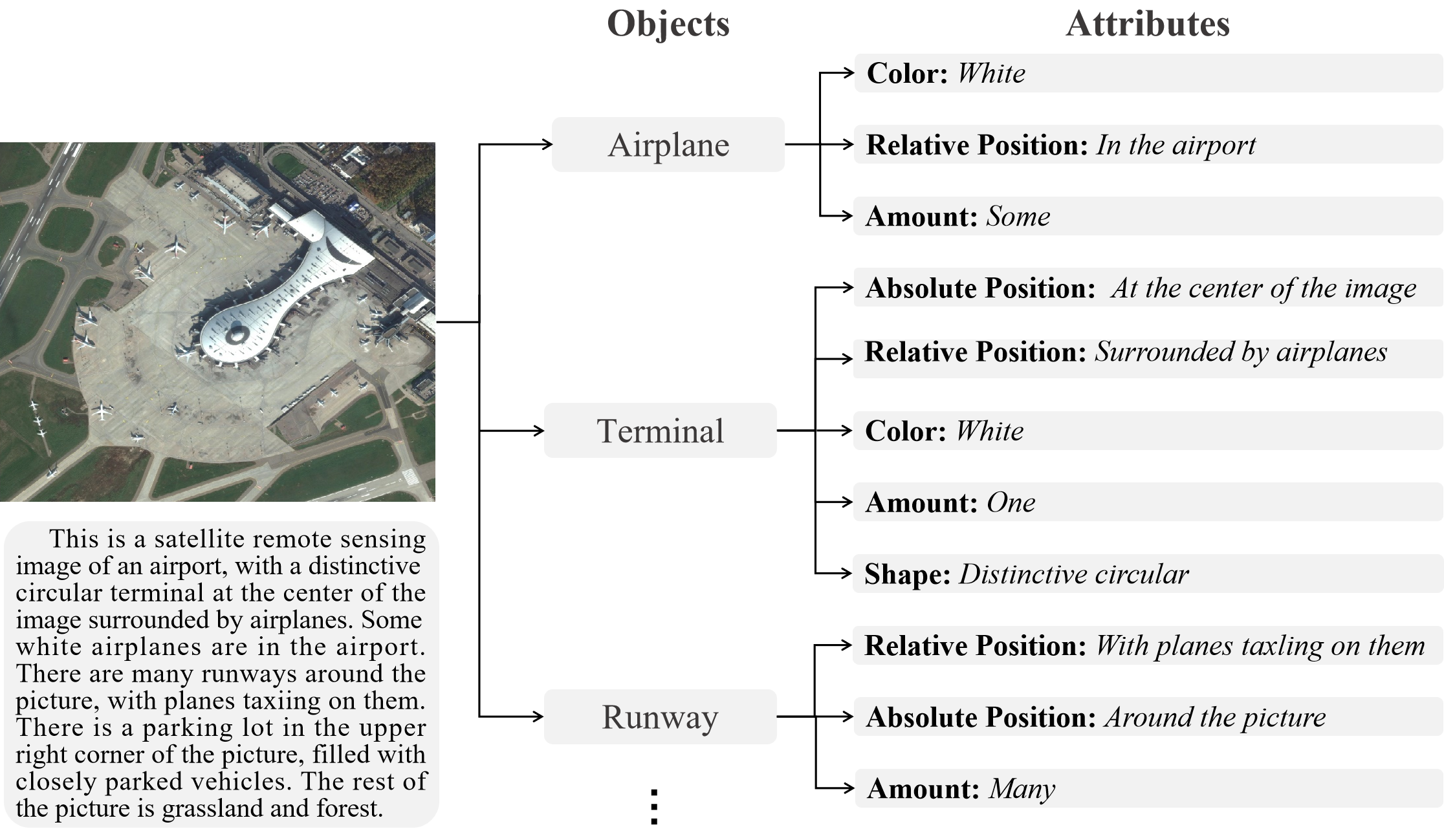}
    \caption{Semantic Unit Tree: The first layer is a virtual root node representing the image. The second layer contains all objects in the image, and the third layer captures the attributes of each object.}
    \label{fig:semantic units}
\end{figure}

\textbf{Minimal Redundancy}: 
Single-round annotation inherently contains no redundancy, as it involves only one annotator. In contrast, under Assumption 2, both CoTalk and parallel annotation involve multiple annotators and rely on LLMs to merge their outputs. While parallel annotation gathers full independent annotations from each annotator, CoTalk captures more refined \textit{"residual"} inputs—focused supplements according to prior contributions. Given that LLMs typically produce output proportional to input length, and that word count serves as a proxy for redundancy (being directly related to entropy ${H(Y)}$, we can compare the overall entropy of the merged outputs. According to Shannon’s estimate of 11.82 bits per English word, higher word counts imply higher redundancy~\cite{GRIGNETTI1964304}. Since CoTalk reduces redundancy through sequential refinement, it achieves lower overall entropy than parallel annotation:
\begin{equation}
	{J}_{\text{min}}^{\text{CoTalk}}(\theta) \approx {J}_{\text{min}}^{\text{Single}}(\theta) > {J}_{\text{min}}^{\text{Par}}(\theta),\label{equation_entropy}
\end{equation}
indicating that CoTalk minimizes redundancy more effectively than parallel annotation, matching the minimal redundancy of single-round annotation while achieving greater information coverage.

\textbf{Human Comprehensibility}: Under Assumption 2, we assume that the merging process is lossless. Since all three annotation methods, CoTalk, parallel, and single-round, rely on human annotators to generate content, their outputs are expected to maintain similar levels of interpretability. Therefore, the human comprehensibility of CoTalk is equivalent to that of the other two methods:
\begin{equation}
	{J}_{\text{int}}^{\text{CoTalk}}(\theta) = {J}_{\text{int}}^{\text{Single}}(\theta) = {J}_{\text{int}}^{\text{Par}}(\theta) \label{equation_human comprehensibility}
\end{equation}

According to the above results, we conclude that CoTalk achieves higher information content, lower redundancy, and comparable human comprehensibility relative to parallel and single-round annotations. Therefore, it offers superior annotation quality, denoted as ${J}_{\theta}^{\text{CoTalk}}$.

\textbf{Theorem 2 (CoTalk boosts efficiency)}: Building on Theorems 1, we establish that CoTalk yields the highest annotation quality. We now compare the efficiency of CoTalk, parallel annotation, and single-round annotation by evaluating the time required to achieve equivalent semantic unit coverage. Since single-round annotation provides significantly lower information sufficiency, we focus on comparing CoTalk and parallel annotation. Assume parallel annotation can match CoTalk’s semantic coverage using ${m}$ rounds, while CoTalk requires only ${n}$ rounds (${m > n}$, as implied by Equation~\ref{equation_info}). The total annotation time is then:
\begin{align}
    {T}_{\text{CoTalk}}^{{n}} &= {n} \cdot {T}_{\text{in}}^{\widetilde{{X}}} + \sum_{{i=1}}^{{n}} ({T}_{\text{in}}^{\widetilde{{Y}}_{\sigma}^{{i-1}}} + {T}_{\text{out}}^{\widetilde{{Y}}^{{i}}}), \nonumber \\
    {T}_{\text{Par}}^{{m}} &= {m} \cdot {T}_{\text{in}}^{\widetilde{{X}}} + \sum_{{i=1}}^{{m}} {T}_{\text{out}}^{\widetilde{{Y}}^{{i}}}
\end{align}
Here, ${T}_{\text{out}}^{\widetilde{{Y}}^{{i}}} = \frac{|{Y}^{{i}}|}{{v}_{\text{out}}}$ denotes the time for producing annotations, and ${T}_{\text{in}}^{\widetilde{{Y}}{\sigma}^{{i-1}}} = \frac{|{Y}{\sigma}^{{i-1}}|}{{v}_{\text{in}}^{\text{text}}}$ represents the time for reviewing prior annotations. The variables ${v}_{\text{out}}$ and ${v}_{\text{in}}^{\text{text}}$ refer to the speech annotation and reading comprehension speeds, respectively. Since ${v}_{\text{out}} > {v}_{\text{in}}^{\text{text}}$~\cite{2016Speech,Brysbaert2019HowMW}, it follows that ${T}_{\text{CoTalk}} < {T}_{\text{Par}}$ when ${n=2}$, as shown in Proof~\ref{app:proof1}. Moreover, the primary time overhead in CoTalk stems from reading prior annotations. However, as semantic coverage increases, the reading time increase per round progressively decreases. In contrast, parallel annotation suffers from growing redundancy, which scales with semantic coverage and leads to increasing time consumption. Hence: ${T}_{\text{CoTalk}}^{n} < {T}_{\text{Par}}^{{m}}$.
demonstrating that CoTalk achieves the same semantic coverage with lower time costs. Consequently, its efficiency surpasses that of both parallel and single-round methods: ${E}_{\text{CoTalk}} > {E}_{\text{Par}} > {E}_{\text{Single}}$.
In summary, CoTalk not only ensures high annotation quality through improved information sufficiency and reduced redundancy, but also maximizes efficiency—minimizing time and labor, and closely approaching \textbf{Pareto Optimality}.

\begin{figure*}
    \centering
    \includegraphics[width=1\linewidth]{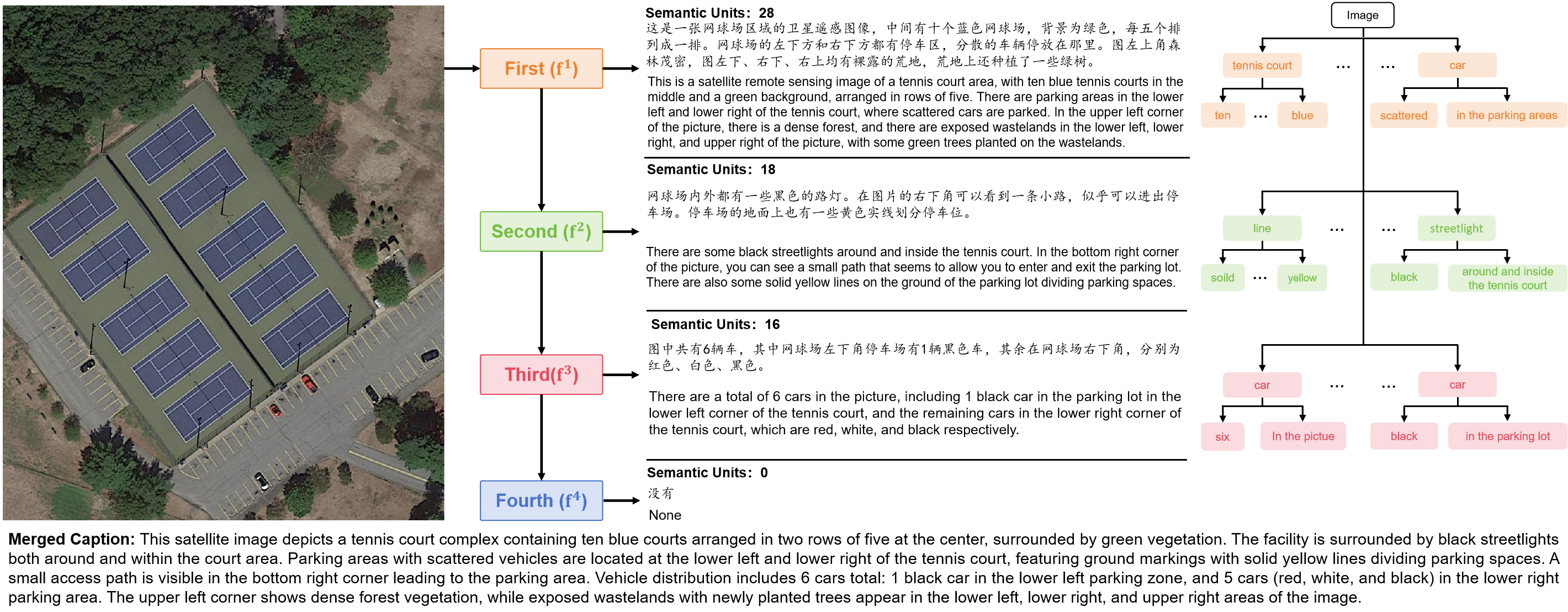}
    \caption{CoTalk Example: The first annotator provides a full image description, while subsequent annotators review and add missing details. As the sequence continues, the number of new semantic units gradually declines.}
    \label{fig:CoTalk}
\end{figure*}

\subsection{CoTalk is Faster in Input and Output}
CoTalk employs a cross-modal interface, using text for input and talking for output, in contrast to traditional single-modality parallel approaches. In this section, we evaluate the efficiency of different input-output modality combinations to determine the most effective configuration.

First, \textbf{talking for output is faster}: The time for talking output can be expressed as: ${T}_{\text{out}}^{\text{talking}}$, and the time for text output can be defined as: ${T}_{\text{out}}^{\text{typing}}$. Existing research shows that talking speed is 161.20 WPM, three times faster than typing, with 20.4\% lower error rates~\cite{2016Speech}, we can directly conclude that ${v}_{\text{out}}^{\text{talking}} > {v}_{\text{out}}^{\text{typing}}$.

Next, \textbf{text for input is faster}: Since eyes spend 10\% to 20\% time rereading during text input, while replaying audio for review is more time consuming compared to rereading text~\cite{Zhan2015ESLRR}. We can conclude that ${v}_{\text{rereading}} > {v}_{\text{relistening}}$. Moreover, we can define the time for listening input as: ${T}_{\text{in}}^{\text{listening}} = {T}_{\text{listening}} + {T}_{\text{relistening}}$, and the time for text input can be expressed as: ${T}_{\text{in}}^{\text{reading}} = {T}_{\text{reading}} + {T}_{\text{rereading}}$. The reading comprehension speed for text is 236 WPM, exceeding the 161 WPM speed to understand talking~\cite{Brysbaert2019HowMW}, which means that ${v}_{\text{reading}} > {v}_{\text{listening}}$. Therefore, we can achieve that ${v}_{\text{in}}^{\text{reading}} > {v}_{\text{in}}^{\text{listening}}$.

From the above results, we conclude that in CoTalk, using multimodal input and output can save more time compared to a single modality approach. Specifically, for annotators' output, talking should be adopted due to its higher speed and accuracy. On the other hand, for understanding previous annotations, text is more effective.

\section{Experiments}
Our experiments aim to validate the effectiveness of different annotation strategies: CoTalk and parallel annotations, as well as to identify the optimal input-output modality for individual annotators, namely talk or text. Finally, we perform an in-depth analysis of the collected human annotation.

\subsection{Evaluation Metrics}
To assess the quality differences among various annotation methods, we conduct a quantitative comparison using both extrinsic and intrinsic metrics. 

\subsubsection{Intrinsic Metric}
Section~\ref{Preliminary} establishes that semantic units serve as a reliable indicator of annotation quality: more units correspond to higher-quality annotations. We further formalize the function ${h}(\cdot)$, which maps annotations to semantic units. Specifically, we utilize LLMs to extract these units and organize them into a hierarchical tree, where the root is a virtual node labeled "Image", the second level contains entities, and subsequent levels represent their associated attributes. The total number of edges quantifies the semantic units. For example, in Figure \ref{fig:semantic units}, ``Terminal'' has five semantic units. 

Building on semantic units, we quantitatively compare CoTalk and parallel annotation in terms of quality and efficiency. As shown in Table~\ref{t3}, two-person CoTalk yields higher-quality annotations (36.72 units/image vs. 33.61) and reduces per-annotator time by 48\%. To assess redundancy, we measure the overlap between outputs from the first and second annotators. Redundancy is defined as the repetition rate of semantic units, including exact matches and semantically similar phrases (\textit{e.g.,} “black car” vs. “black vehicle”) exceeding a similarity threshold using Sentence-BERT~\cite{Reimers2019SentenceBERTSE}. CoTalk shows lower redundancy (29.76\%) compared to parallel annotation (69.12\%), highlighting the inefficiencies of independent annotation without shared context. 

\subsubsection{Extrinsic Metric}
Extrinsic Metric assesses the practical effectiveness of annotation methods. We focus on image-text retrieval, a task closely aligned with the goal of maximizing mutual information in InfoNCE. In information theory, higher mutual information indicates better cross-modal predictability. Therefore, stronger retrieval performance, reflects more informative and semantically aligned annotations. To evaluate this, we fine-tune Long-CLIP~\cite{Zhang2024LongCLIPUT} using annotations generated by CoTalk and parallel methods, then test the models on the remote sensing benchmarks: RSICD~\cite{Lu2017ExploringMA}, RSITMD~\cite{Yuan2021ExploringAF}, and UCM-Captions~\cite{7546397}. As shown in Table~\ref{t2}, the model trained with CoTalk annotations achieves the highest average retrieval score (41.13\%), outperforming the model trained with parallel annotations (40.52\%). More details are in~\ref{app:Extrinsic Metric}.

\begin{table}[t]
\centering
\caption{Information on different methods. Redundancy refers to internal components within each annotation method.}
\renewcommand\arraystretch{1.25}
\resizebox{0.5\textwidth}{!}{
\begin{tabular}{c|cccc}
    \hline
        Method & \#Semantic Units & Time & Speed(units/s) & Duplication $\downarrow$ \\ \hline
        Parallel & 33.61 & 118.30& 0.30 & 69.12\\ 
        \hline
        \rowcolor{blue!10}
        \textbf{CoTalk}  & \textbf{36.72} & \textbf{88.43}& \textbf{0.42} & \textbf{29.76}  \\ \hline
    \end{tabular}
}
\label{t3}
\end{table}

\begin{table}[t]
\centering
\caption{Comparison of quality and efficiency between speaking and typing, and comprehension between listening and reading.}
\renewcommand\arraystretch{1.05}
\resizebox{0.48\textwidth}{!}{
\begin{tabular}{c|cccc}
\hline
        Method & \# Semantic Units & Time & Speed(units/s) \\ \hline
        Talking & 46.65 & 116.65 & 0.40 \\ 
        Typing & 33.15 & 199.15 & 0.17 \\ \hline
        ~ & Accuracy & Time  & \# Frequency \\ \hline
        Listening & 94.00 & 70.20 & \multirow{2}{*}{2.22} \\ 
        Reading & 100.00 & 55.20  \\ \hline
    \end{tabular}
}
\label{t4}
\end{table}

With intrinsic and extrinsic metrics, CoTalk annotation consistently outperforms parallel annotation in quality, efficiency, and redundancy.

\subsection{Qualitative Analysis} 
From a qualitative perspective, we explore the concept of \textit{``residual''} within the CoTalk annotation. Annotators typically focus on entities overlooked by previous annotators, followed by missing attributes of these entities, as illustrated in Figure \ref{fig:CoTalk}. These attributes, such as color, position (relative and absolute), quantity, shape, and size, align with our defined semantic units. Notably, annotators rarely correct prior annotations, probably because of the high accuracy of initial inputs. Instead, each annotation round adds new entities or attributes, embodying the concept of \textit{``residual''} in CoTalk and enriching the description without redundancy.

\subsection{Talking is Faster for Output, Reading is Faster for Input}
After establishing CoTalk as an effective framework for multi-annotator collaboration, we evaluate which input and output modalities best maximize annotator efficiency, comparing speech and text in both input (viewing prior annotations) and output (producing new annotations) modes.

\begin{table}[t]
\centering
\caption{Validate annotation quality by evaluating performance in downstream retrieval tasks.}
\renewcommand\arraystretch{1.25}
\resizebox{0.5\textwidth}{!}{
\begin{tabular}{c|cccc}

\hline
        Method & RSICD & RSITMD & UCM-Captions &  Average $\uparrow$\\  \hline 
        Zero-shot & 21.24 & 31.10  & 65.01 & 37.94\\ 
        Parallel & 22.57±0.06 & \textbf{33.97±0.05} & 65.01±0.07 & 40.52 \\ 
        \hline
        \rowcolor{blue!10}
        \textbf{CoTalk} & \textbf{23.63 ± 0.05} & 33.83 ± 0.09 & \textbf{65.94 ± 0.06} & \textbf{41.13} \\ \hline
    \end{tabular}
}
\label{t2}
\end{table}

\begin{table}[t]
\centering
\caption{The Relationship Between Annotation Quality and Annotation Cycles: Annotation quality is analyzed in relation to annotation cycles, where each cycle corresponds to approximately 10 minutes of annotation.}
\renewcommand\arraystretch{1.20}
\resizebox{0.48\textwidth}{!}{
\begin{tabular}{c|cccc}
    \hline
        Cycle & \# Amount & \# Semantic units &  Time & Speed(units/s)  \\ \hline
        1 & 4 & 104 & 745.14 & 0.14  \\ 
        2 & 8 & 289 & 628.26 & 0.46  \\ 
        3 & 9 & 308 & 770.00 & 0.40  \\ 
        4 & 6 & 180 & 720.00 & 0.25  \\ \hline
    \end{tabular}
}
\label{t7}
\end{table}

\subsubsection{Annotation Output: Talk vs. Type}
To compare the quality and efficiency of talking versus text annotation, we recruit eight experienced annotators, divided into two groups of four. Within each group, two annotators leverage talk input, while the other two use text input via keyboard to annotate the same set of images. The results are averaged across the annotation methods for comparison. As shown in Table \ref{t4}, talking yields an average of 13.50 more units per image than typing, suggesting greater detail and completeness. Additionally, talking requires 116.65 seconds per image, 41\% faster than the 199.15 seconds needed for typing, demonstrating a clear efficiency advantage.

\subsubsection{Annotation Input: Read vs. Listen} 
To compare comprehension from speech and text, we perform an experiment using various images and LLM-generated questions according to human annotations. Eight annotators answer five questions per image in both text and audio formats. As Table \ref{t4} shows, reading is on average 15 seconds faster and 6.38\% more accurate. Listeners need 2.2 replays on average for detailed information, while rereading text takes less time. The gap widens in complex scenes (\textit{e.g.,} urban or industrial).

\begin{figure}
    \centering
    \includegraphics[width=1\linewidth]{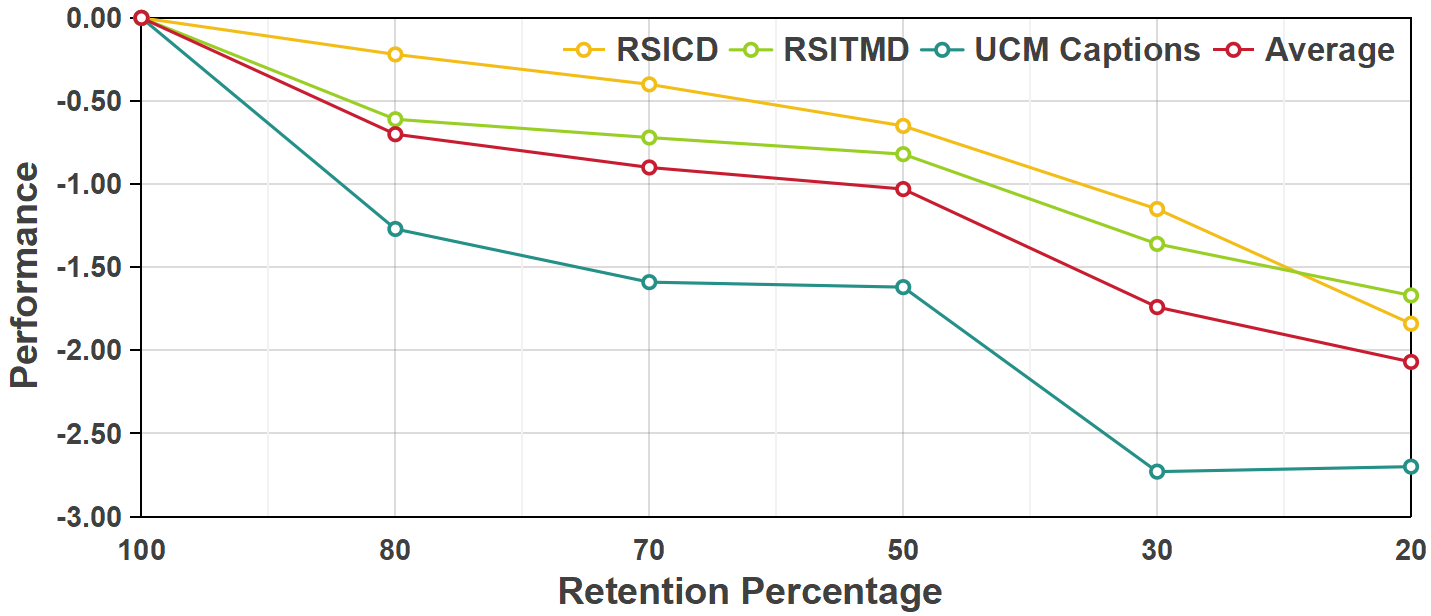}
    \caption{Consistency Analysis of Intrinsic Metric semantic units and Extrinsic Metric by finetuning the Long-CLIP with fixed ratio data reduced. Each point represents its score minus the score from the full dataset.}
    \label{fig:Extrinsic}
\end{figure}


These findings suggest that relying solely on speech (talking) or text for both input and output is suboptimal. Instead, a hybrid approach using text for input and speech for output can achieve both higher annotation quality and greater efficiency.

\subsection{Further Analysis}

\subsubsection{Consistency between Extrinsic and Intrinsic Metrics}
Given the strength of extrinsic metrics in evaluating annotation quality, we test whether intrinsic metrics, specifically semantic unit count, serve as reliable proxies. We fine-tune Long CLIP~\cite{Zhang2024LongCLIPUT} on full CoTalk annotations, then reduce semantic units per image by fixed ratios and evaluate on remote sensing retrieval benchmarks. As shown in Figure~\ref{fig:Extrinsic}, performance decreases steadily with fewer units, averaging a 2.07\% decline when only 20\% remain. This trend confirms that the amount of semantic units directly impacts annotation informativeness and model performance, which is consistent with extrinsic metric. The results also underscore the value of dense captions for robust vision-language alignment. More details are in~\ref{app: Consistency}.

\label{Consistency between Extrinsic and Intrinsic Metrics:}
\subsubsection{Annotator Ability Over Time}
We recruit untrained annotators for CoTalk and analyze their initial annotation cycles (10 minutes each) to gauge ability and learning curves. As shown in Table~\ref{t7}, novices match experienced speeds (0.46 units/s) after one session, suggesting rapid adaptation via previous examples and familiarity with the task. However, the speed drops from 0.40 to 0.25 units in round four, indicating fatigue or reduced efficiency over time, which highlights the importance of considering session duration and rest intervals in future workflow designs.

\begin{figure}
    \centering
    \includegraphics[width=1\linewidth]{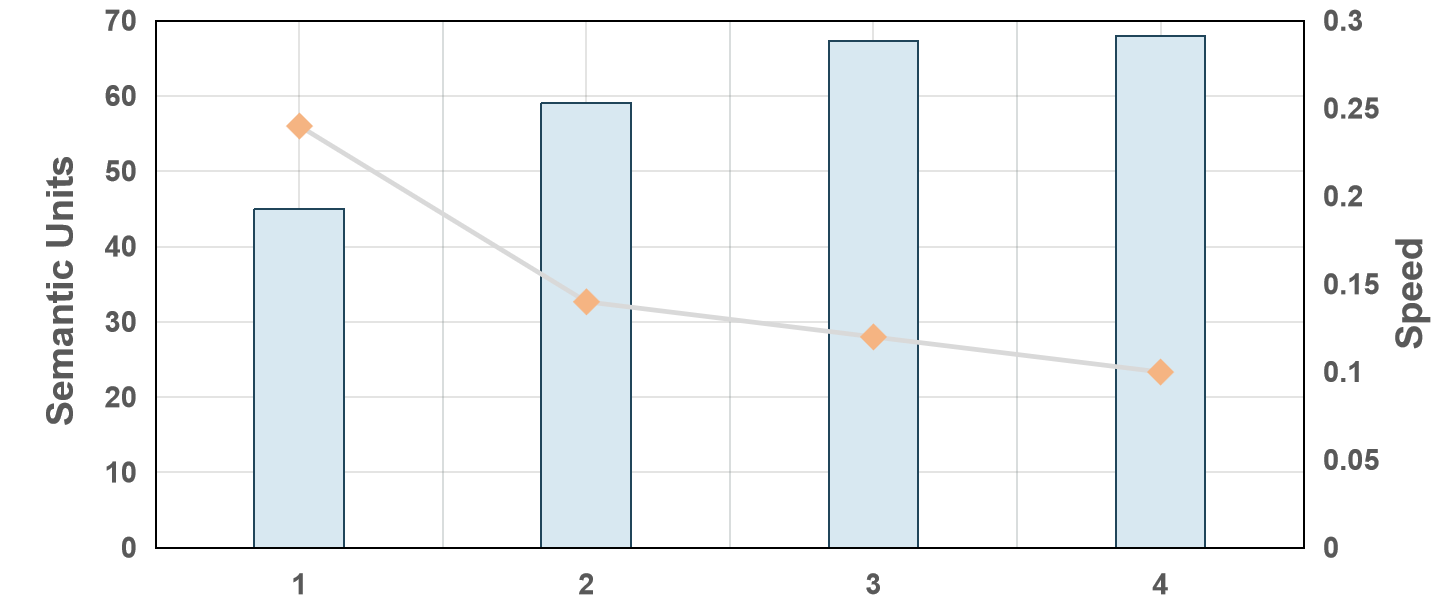}
    \caption{The relationship between images with varying semantic unit counts and the number of required annotators (yellow), and the change in annotation speed at different positions in CoTalk (blue).}
    \label{fig:f6}
\end{figure}

\subsubsection{Full Image Annotation: Annotator Count and Speed}
Multiple annotators sequentially annotate each image using CoTalk until the last annotator deems it sufficiently detailed. We record annotation speed and the number of annotators per image. As shown in Figure~\ref{fig:f6} (blue), speed decreases as early annotators cover obvious content, leaving finer details for later ones. On average, 3.2 annotators are enough per image (Figure~\ref{fig:f6}, yellow). Moreover, images with more units need more annotators, making semantic unit count a strong complexity indicator. For images with 60 or fewer units, two annotators suffice; more complex ones need at least three.


Given these findings, intrinsic and extrinsic metrics align, since fewer semantic units lead to lower retrieval performance. Furthermore, novice annotators in CoTalk quickly perform as well as experienced ones. Although annotation speed slows over iterations, only 3.2 annotators are needed on average to produce a sufficiently detailed annotation.

\section{Conclusion}
Our work presented two key insights: sequential annotation minimized redundancy compared to parallel approaches, and humans achieved higher efficiency by reading text input and talking output. Building on these, we proposed Chain-of-Talkers (CoTalk), a cross-modal, sequential collaboration framework that enhanced human annotation for image captioning. Both theoretical and empirical analysis confirmed that CoTalk significantly improved caption quality within fixed budget constraints.

\section*{Limitations}
Although CoTalk has proven effective in enhancing image caption quality and efficiency, several limitations remain worth discussing.

\textbf{Assumptions on Human Ability.} Our comparison of information sufficiency, human comprehensibility, and efficiency between sequential annotation and CoTalk assumes consistent annotator abilities. However, in practice, annotator capabilities often vary. In sequential annotation, later annotators may identify more \textit{residual} information than earlier ones, while in parallel annotation, annotators may differ significantly in annotation time and quality for the same image. Additionally, we assume all annotations are accurate, but in reality, factors such as annotator ability and attention may cause deviations between the annotated semantic units and the true content of the image.

\textbf{Assumptions on Model Ability.} CoTalk relies on large models to merge annotations from multiple annotators. While our theoretical analysis assumes that these models can accurately consolidate diverse inputs without information loss, practical challenges may arise—particularly when annotators provide contradictory descriptions or inconsistent expressions for the same object. Additionally, CoTalk employs a speech-to-text model to transcribe annotator speech. Although state-of-the-art models such as~\cite{An2024FunAudioLLMVU, Radford2022RobustSR} achieve high accuracy, transcription errors can still occur, especially with unclear or imprecise speech. Further exploration and careful selection of robust language and speech-to-text models are critical to improving CoTalk’s overall reliability.

\section*{Ethics Statement}
In developing and evaluating the Chain-of-Talkers (CoTalk) framework for image captioning annotation, we upheld rigorous ethical standards to ensure integrity, fairness, and accountability throughout the research process.

\textbf{Participant Welfare and Informed Consent}:
All human annotators were fully informed about the nature, purpose, and procedures of the annotation tasks. Participation was entirely voluntary, and written informed consent was obtained from each individual. Annotators were informed of how their data would be utilized for research purposes. We ensured the privacy and anonymity of all participants by removing personal identifiers from the data and results.

\textbf{Data Integrity and Transparency}: 
We maintained strict data integrity, accurately recording annotation data without manipulation or fabrication. Secure data management practices were implemented to prevent data loss or unauthorized access. All relevant data, results, and methodologies are reported transparently to enable reproducibility and facilitate verification by other researchers.

\textbf{Broader Impact}: 
The CoTalk framework enhances annotation quality and efficiency, benefiting downstream tasks such as image retrieval and visual understanding. However, annotations may still reflect annotator biases or image-related imbalances, potentially influencing model behavior. High-quality annotations can also be misused in sensitive domains like surveillance. We urge responsible use of CoTalk, with attention to fairness, privacy, and application-specific risks.

\bibliography{ref.bib}

\appendix
\section*{\centering\textbf{Appendix}}

\section{Related Work}

A growing trend in image captioning research was the emphasis on generating more comprehensive captions~\cite{Cho2025PerceptionLMOD, Bolya2025PerceptionET, Shabbir2025GeoPixelPG, Hua2024FINECAPTIONCI, Athar2024ViCaSAD, Singla2024FromPT, Ma2024GromaLV, Nguyen2023ImprovingMD}, aiming for higher information sufficiency through broader semantic coverage. Recently, numerous dense captioning datasets emerged, including high-quality manually annotated datasets~\cite{Onoe2024DOCCIDO, Deitke2024MolmoAP, Hu2023RSGPTAR} and pseudo-labeled datasets generated by models~\cite{Ge2024VisualFC, Singla2024FromPT, Chen2023ShareGPT4VIL, Ou2025GeoPixML}. Although some approaches incorporated prior knowledge~\cite{Ou2025GeoPixML, Li2024LHRSBotNovaIM} or utilized visual models for secondary correction~\cite{Ge2024VisualFC, Li2023MonkeyIR}, the quality of model-generated detailed captions remained notably inferior to that of human annotations.

However, traditional human annotation was labor-intensive, and the associated time and coordination costs significantly constrained scalability and efficiency. While several studies demonstrated that involving multiple annotators improved the detail and accuracy of descriptions~\cite{Hu2023RSGPTAR, Onoe2024DOCCIDO}, these benefits came at the expense of a higher annotation overhead. To improve annotation efficiency, some studies introduced models to assist human annotators. For example, models could first identify key objects or generate preliminary descriptions, which were then refined by humans~\cite{Cho2025PerceptionLMOD, Garg2024ImageInWordsUH}. However, these approaches still relied on traditional typing for annotation, which remained time-consuming. More recently, several works explored using speech input to accelerate annotation, reducing typing time costs and mitigating issues such as annotator manipulation during large-model-assisted labeling~\cite{Deitke2024MolmoAP, Athar2024ViCaSAD}. Nonetheless, most of these methods relied on merging multi-person speech annotations, which often introduced redundancy and partially offset the efficiency gains.

To overcome these limitations, we proposed CoTalk, a human-AI collaborative annotation framework based on two key insights. First, sequential annotation, where subsequent annotators provided \textit{``residual''} supplements to previous annotations, yielded higher quality and efficiency than parallel annotation. Second, combining text-based input with talk-based output in this sequential setting optimized both speed and accuracy, outperforming traditional single-modality approaches.

\begin{table*}[t]
\centering
\caption{The simulation results of sequence and parallel on large model}
\addtolength{\tabcolsep}{-3pt}
\renewcommand\arraystretch{2.00}
\resizebox{1.00\textwidth}{!}{
\begin{tabular}{ccccccccccccccccccccccccc}
\hline
        \multirow{3}{*}{Model} & \multicolumn{8}{c}{RSICD}  & \multicolumn{8}{c}{RSITMD} & \multicolumn{8}{c}{UCM-Captions} \\ 
        \cmidrule(lr){2-9} \cmidrule(lr){10-17} \cmidrule(lr){18-25}
        ~ & \multicolumn{3}{c}{Image to Text} & \multirow{2}{*}{Average}  & \multicolumn{3}{c}{Text to Image} & \multirow{2}{*}{Average}  & \multicolumn{3}{c}{Image to Text} & \multirow{2}{*}{Average}  & \multicolumn{3}{c}{Text to Image} & \multirow{2}{*}{Average}& \multicolumn{3}{c}{Image to Text} & \multirow{2}{*}{Average}  & \multicolumn{3}{c}{Text to Image} & \multirow{2}{*}{Average}  \\ 
        \cmidrule(lr){2-4} \cmidrule(lr){6-8} \cmidrule(lr){10-12} \cmidrule(lr){14-16} \cmidrule(lr){18-20} \cmidrule(lr){22-24} 
        ~ & R@1 & R@5 & R@10 &~ & R@1 & R@5 & R@10 & ~ & R@1 & R@5 & R@10 & ~ & R@1 & R@5 & R@10 & ~ & R@1 & R@5 & R@10 & ~ & R@1 & R@5 & R@10 &  \\ \hline
        Zero-shot & 9.70  & 23.88  & 38.15  & 23.91  & 5.13  & 19.11  & 29.85  & 18.03   & 13.50  & 33.63  & 45.58  & 30.90  & 10.52  & 31.57  & 47.10  & 29.73   & 39.50  & 80.00  & 90.48  & 69.99  & 28.38  & 59.68  & 77.19  & 55.08   \\ \hline
        Parallel & 10.06  & \textbf{25.62}  & 37.69  & 24.46  & 6.21  & 20.60  & 32.84  & 19.88   & \textbf{15.93}  & 36.73  & 47.57  & 33.41  & 12.27  & 35.58  & 50.31  & 32.72   & \textbf{41.43}  & \textbf{79.52}  & 91.43  & \textbf{70.79}  & 30.24  & \textbf{64.19}  & \textbf{82.76}  & 59.06   \\ \hline
        \rowcolor{blue!10}
        CoTalk & \textbf{10.25}  & 25.53  & \textbf{38.33}  & \textbf{24.70}  & \textbf{6.78}  & \textbf{21.82}  & \textbf{33.97}  & \textbf{20.86}   & 15.49  & \textbf{38.05}  & \textbf{49.78}  & \textbf{34.44}  & \textbf{12.88}  & \textbf{36.86}  & \textbf{52.05}  & \textbf{33.93}   & 40.00  & \textbf{79.52}  & \textbf{91.90}  & 70.47  & \textbf{31.56}  & \textbf{64.19}  & 82.23  & \textbf{59.33}  \\ \hline
    \end{tabular}
}
\label{app:simulation}
\end{table*}
\section{Proof for Theorem}
\textbf{Proof 1} When the number of annotators in CoTalk is ${n = 2}$, under the same semantic unit coverage, the total time for parallel annotation exceeds that of CoTalk annotation, indicating that CoTalk is more efficient.
Before proceeding, we review the necessary assumptions:  
(1) Taggers have equal abilities, therefore the number of semantic units supplemented by the second annotator $\widetilde{{Y}}_{\text{CoTalk}}^{2}$ is less than the number initially annotated by the first $\widetilde{{Y}}_{\text{CoTalk}}^{1}$;  
(2) According to Equation~\ref{equation_info}, the number of annotators in parallel annotation ${m}$ is greater than in CoTalk ${n}$, i.e., ${( m > n )}$, with both ${m}$ and ${n}$ as integers;  
(3) Prior research shows that the output speed for voice annotation ${v}_{\text{out}}$ is faster than the reading comprehension speed ${v}_{\text{in}}^{\text{text}}$~\cite{2016Speech, Brysbaert2019HowMW}. Given ${n=2}$, the CoTalk annotation time is:
\begin{equation}
    {T}_{\text{CoTalk}}^{2} = {2} \cdot {T}_{\text{in}}^{\widetilde{{X}}} + {T}_{\text{in}}^{\widetilde{{Y}}_{\text{CoTalk}}^{1}} + \sum_{{i=1}}^{2} {T}_{\text{out}}^{\widetilde{{Y}}_{\text{CoTalk}}^{{i}}}
\end{equation}
For parallel annotation:
\begin{equation}
    {T}_{\text{Par}}^{m} = {m}\cdot{T}_{\text{in}}^{\widetilde{{X}}} + \sum_{{i=1}}^{{m}} {T}_{\text{out}}^{\widetilde{{Y}}_{\text{Par}}^{{i}}}
\end{equation}
where ${m > n}$. Considering an extreme case where ${m=3}$ means that the number of effective semantic units completed by three parallel annotators is the same as the two annotators in CoTalk, and assuming (according to premise 1) that all annotators have the same ability, each parallel annotator completes ${{Y}}_{\text{par}}^{1}$ semantic units. Therefore, the time for parallel annotation becomes:
\begin{equation}
    {T}_{\text{Par}}^{3} = {3}\cdot{T}_{\text{in}}^{\widetilde{{X}}} + {3}\cdot{T}_{\text{out}}^{\widetilde{{Y}}_{\text{Par}}^{1}}
\end{equation}
The time difference between parallel and CoTalk annotation for the same number of semantic units is:
\begin{equation}
\Delta {T(n=2, m=3)} = {T}_{\text{Par}}^{3} - {T}_{\text{CoTalk}}^{2}
\end{equation}
Expanding:
\begin{equation}
    \Delta {T} = {T}_{\text{in}}^{\widetilde{{X}}} + {2}\cdot{T}_{\text{out}}^{\widetilde{{Y}}_{\text{Par}}^{1}} - {T}_{\text{out}}^{\widetilde{{Y}}_{\text{Cotalk}}^{{2}}} - {T}_{\text{in}}^{\widetilde{{Y}}_{\text{CoTalk}}^{1}} 
\end{equation}
\begin{equation}
    \Delta {T} = {T}_{\text{in}}^{\widetilde{{X}}} + \frac{{2}\cdot|\widetilde{{Y}}_{\text{Par}}^{1}| - |\widetilde{{Y}}_{\text{CoTalk}}^{2}|}{{v}_{\text{out}}} - \frac{|\widetilde{{Y}}_{\text{CoTalk}}^{1}|}{{v}_{\text{in}}^{\text{text}}}
\end{equation}
Since ${v}_{\text{out}} < {v}_{\text{in}}^{\text{text}}$, it follows that:
\begin{equation}
    \frac{\widetilde{{Y}}_{\text{CoTalk}}^{1}}{{v}_{\text{in}}^{\text{text}}} < \frac{\widetilde{{Y}}_{\text{CoTalk}}^{1}}{{v}_{\text{out}}}
\end{equation}
thus:
\begin{equation}
    \Delta {T} > {T}_{\text{in}}^{\widetilde{X}} + \frac{\widetilde{{Y}}_{\text{CoTalk}}^{1} - \widetilde{{Y}}_{\text{CoTalk}}^{2}}{{v}_{\text{out}}} > 0
\end{equation}
Given that taggers have identical abilities $\widetilde{{Y}}_{\text{par}}^{1} = \tilde{{Y}}_{\text{CoTalk}}^{1}$ and $\widetilde{{Y}}_{\text{CoTalk}}^{1} > \widetilde{{Y}}_{\text{CoTalk}}^{2}$, the positive time difference $\Delta {T}$ is established.
Moreover, as ${m}$ increases, the total time for parallel annotation grows. Therefore, for any ${m > n}$, the time difference satisfies:
\begin{equation}
    \Delta {T(n=2, m>n)} = {T}_{\text{Par}}^{m} - {T}_{\text{CoTalk}}^{2} > 0
\end{equation}
confirming that CoTalk annotation remains more efficient than parallel annotation under these base conditions.
\label{app:proof1}

\section{Support for Assumption}
\begin{figure}[h]
    \centering
    \includegraphics[width=1\linewidth]{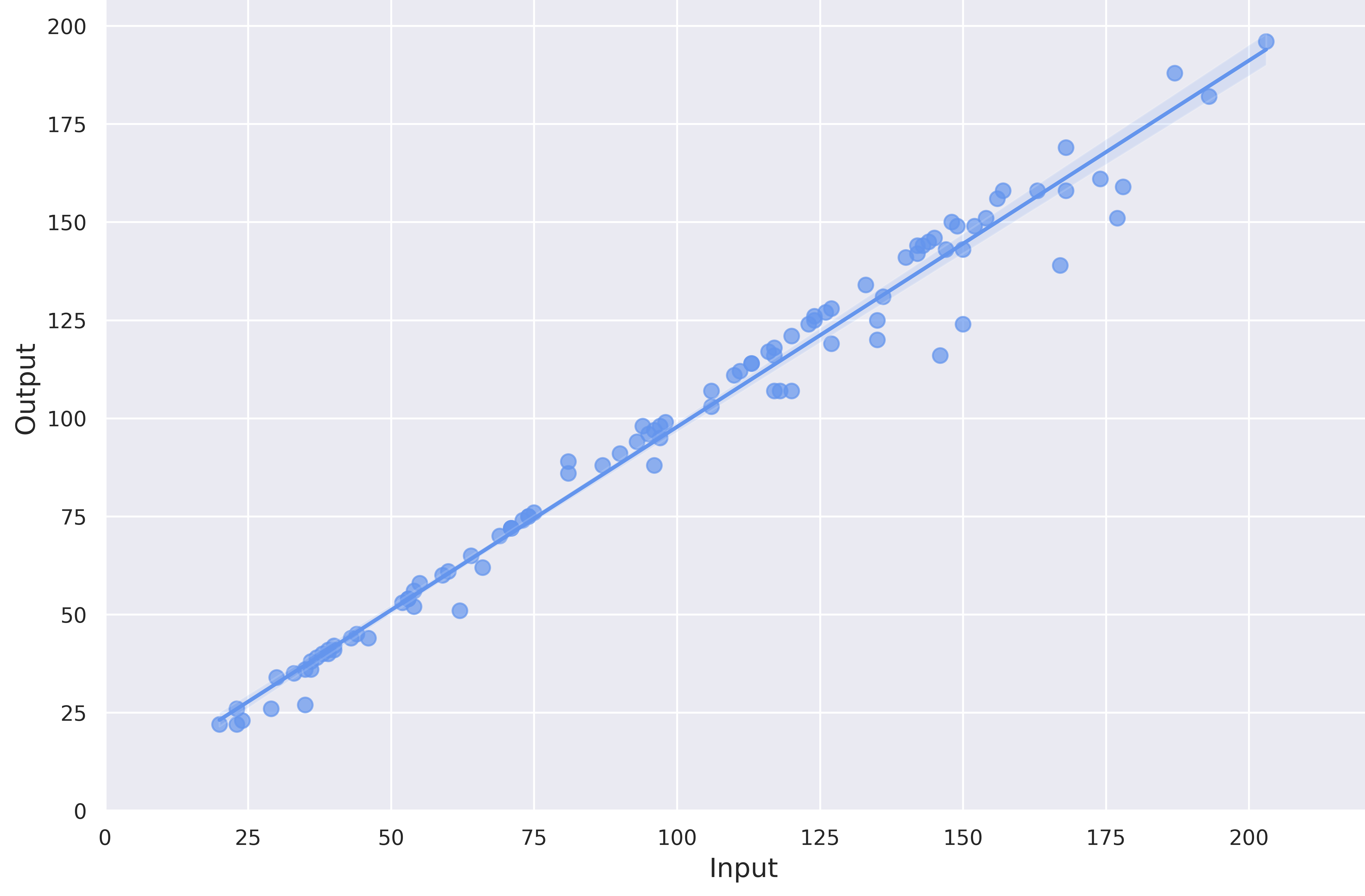}
    \caption{Proof of Assumption 2: The Merging Function Demonstrates a Positive Input-Output Correlation.}
    \label{fig:support1}
\end{figure}

\textbf{Support for LLM input-output consistency}: To validate the consistency of $\sigma(\cdot)$, specifically, that more input yields more output. We merge multi-person annotations at the sentence level. Token counts exclude prompt tokens, considering only those from the annotated sentences to be merged. As shown in Figure~\ref{fig:support1}, output tokens increase with input tokens, which is consistent with the hypothesis.
\label{app:support1}

\begin{table*}[t]
\centering
\caption{The complete experimental results of Extrinsic Metric in the Ret-3 dataset benchmark evaluation.}
\addtolength{\tabcolsep}{-3pt}
\renewcommand\arraystretch{1.70}
\resizebox{1.00\textwidth}{!}{
\begin{tabular}{cccccccccccccccccccccc}
\toprule
\hline
        \multirow{3}{*}{Model} & \multicolumn{7}{c}{RSICD}  & \multicolumn{7}{c}{RSITMD} & \multicolumn{7}{c}{UCM-Captions} \\ 
        \cmidrule(lr){2-8} \cmidrule(lr){9-15} \cmidrule(lr){16-22}
        ~ & \multicolumn{3}{c}{Image to Text} & \multicolumn{3}{c}{Text to Image} & \multirow{2}{*}{Average}  & \multicolumn{3}{c}{Image to Text} & \multicolumn{3}{c}{Text to Image} & \multirow{2}{*}{Average}& \multicolumn{3}{c}{Image to Text} & \multicolumn{3}{c}{Text to Image} & \multirow{2}{*}{Average}  \\ 
        \cmidrule(lr){2-4} \cmidrule(lr){5-7} \cmidrule(lr){9-11} \cmidrule(lr){12-14} \cmidrule(lr){16-18} \cmidrule(lr){19-21} 
        ~ & R@1 & R@5 & R@10 & R@1 & R@5 & R@10 & ~ & R@1 & R@5 & R@10 & R@1 & R@5 & R@10 & ~ & R@1 & R@5 & R@10 & R@1 & R@5 & R@10 &  \\ \hline
        Zero-shot & 9.70  & 23.88  & 38.15  & 5.13  & 19.11  & 29.85  & 20.97  & 13.50  & 33.63  & 45.58  & 10.52  & 31.57  & 47.10  & 30.32 & 39.50  & 80.00  & 90.48  & 28.38  & 59.68  & 77.19  & 62.54 \\ 
        Parallel & 9.79 & 26.19 & 38.46 & 6.19 & 20.96 & 33.83 & 22.57  & \textbf{16.24} & \textbf{36.73} & \textbf{48.76} & \textbf{13.52} & 36.61 & 52.02 & \textbf{33.97}  & 44.57 & 76.48 & 90.10 & 32.26 & 64.83 & 81.80 & 65.01 \\ \hline
        \rowcolor{blue!10}
        CoTalk & \textbf{10.41} & \textbf{27.32} & \textbf{39.65} & \textbf{6.49} & \textbf{22.5} & \textbf{35.38} & \textbf{23.63} & 15.49 & 35.62 & 48.27 & 13.17 & \textbf{37.35} & \textbf{53.04} & 33.83 & \textbf{46.67} & \textbf{78.38} & 89.24 & 32.41 & 64.77 & \textbf{84.19} & \textbf{65.94} \\ \hline
    \end{tabular}
}
\label{complete_t1}
\end{table*}

\begin{table*}[htbp]
\centering
\caption{Consistency analysis of Intrinsic Metrics Complete data results.}
\addtolength{\tabcolsep}{-3pt}
\renewcommand\arraystretch{1.70}
\resizebox{1.00\textwidth}{!}{
\begin{tabular}{cccccccccccccccccccccc}
\toprule
\hline
        \multirow{3}{*}{Percentage} & \multicolumn{7}{c}{RSICD}  & \multicolumn{7}{c}{RSITMD} & \multicolumn{7}{c}{UCM-Captions} \\ 
        \cmidrule(lr){2-8} \cmidrule(lr){9-15} \cmidrule(lr){16-22}
        ~ & \multicolumn{3}{c}{Image to Text} & \multicolumn{3}{c}{Text to Image} & \multirow{2}{*}{Average}  & \multicolumn{3}{c}{Image to Text} & \multicolumn{3}{c}{Text to Image} & \multirow{2}{*}{Average}& \multicolumn{3}{c}{Image to Text} & \multicolumn{3}{c}{Text to Image} & \multirow{2}{*}{Average}  \\ 
        \cmidrule(lr){2-4} \cmidrule(lr){5-7} \cmidrule(lr){9-11} \cmidrule(lr){12-14} \cmidrule(lr){16-18} \cmidrule(lr){19-21} 
        ~ & R@1 & R@5 & R@10 & R@1 & R@5 & R@10 & ~ & R@1 & R@5 & R@10 & R@1 & R@5 & R@10 & ~ & R@1 & R@5 & R@10 & R@1 & R@5 & R@10 &  \\ \hline
        100\%  & 11.25 & 27.45 & 40.35 & 6.42 & 21.08 & 33.69 & 23.38 & 17.04 & 35.18 & 49.12 & 13.45 & 36.24 & 51.63 & 33.78 & 43.33 & 79.52 & 91.9 & 34.75 & 63.13 & 81.17 & 65.63  \\ 
        80\% & 11.80 & 27.17 & 40.53 & 6.26 & 20.64 & 32.58 & 23.17 & 17.48 & 34.73 & 47.57 & 12.69 & 36.10 & 50.45 & 33.17  & 41.9 & 77.62 & 90.48 & 33.16 & 61.27 & 81.70 & 64.36 \\
        70\% & 11.34 & 26.53 & 40.62 & 6.19 & 20.51 & 32.70 & 22.98  & 16.37 & 34.51 & 46.90 & 12.65 & 35.91 & 52.01 & 33.06 & 43.81 & 77.62 & 90.95 & 30.50 & 59.68 & 81.70 & 64.04  \\ 
        50\% & 10.61 & 27.17 & 40.71 & 5.79 & 19.87 & 32.20 & 22.73 & 16.59 & 34.73 & 47.57 & 12.51 & 35.35 & 51.01 & 32.96  & 40.48 & 78.57 & 90.48 & 31.03 & 62.60 & 80.90 & 64.01  \\ 
        30\% & 10.43 & 27.08 & 40.44 & 5.47 & 19.01 & 30.97 & 22.23  & 15.49 & 34.73 & 46.90 & 11.18 & 35.06 & 51.20 & 32.43 & 38.10 & 78.57 & 90.95 & 28.91 & 61.01 & 79.84 & 62.90  \\ 
        20\% & 10.16 & 25.62 & 39.98 & 4.75 & 18.38 & 30.35 & 21.54 & 15.27 & 34.29 & 46.68 & 10.05 & 35.49 & 50.87 & 32.11 & 41.43 & 75.71 & 88.57 & 30.50 & 61.01 & 80.37 & 62.93 \\ \hline
    \end{tabular}
}
\label{complete_t2}
\end{table*}

\section{Simulation}
Due to the high cost of manual annotation, we utilize only 427 samples from the full CoTalk dataset for our experiments. To supplement this, we simulate data generation using LLMs. Specifically, we apply the model to the entire DOTA training set, approximately three times larger than the original CoTalk dataset, to generate annotations following both the CoTalk and parallel annotation methods, as shown in Table \ref{t1_2}. We then fine-tune the Long-CLIP-L model on the resulting datasets using three V100 GPUs, with a batch size of 16, a learning rate of 1e-6, and for 4 epochs.

We evaluate the models on remote sensing retrieval tasks, with the results summarized in Table~\ref{app:simulation}. CoTalk achieves an average score of 40.62\% across six tasks and three datasets, outperforming both the parallel method (40.05\%) and the zero-shot baseline (37.64\%). Notably, CoTalk surpasses both baselines in almost every task.

\section{Complete Experimental Results}
The comprehensive experimental results, including both internal and external metrics, for the image-text retrieval downstream task evaluated on the Ret-3 dataset are summarized in table \ref{complete_t1} and table \ref{complete_t2}.

\section{Prompt}
\subsection{Merge Annotation}
\label{Merge Annotation}
Both CoTalk and parallel annotation require a large model to integrate individual annotations. This section provides a detailed analysis of the merging process. With the instructions in Table \ref{t10} and Table \ref{t11}, we design specific prompts to guide the LLMs in merging image captions from both annotation methods, ultimately producing semantically rich descriptions.

\subsection{Comparison of Talk and Text in Annotation Input}
To verify that reading comprehension is faster than listening comprehension, we utilize a text-based LLM to generate questions according to fundamental facts. Participants answer these questions under both reading and listening conditions. The prompts, detailed in Table \ref{t14}, are designed to generate high-quality questions with varying levels of difficulty, gradually incorporating finer-grained image content to ensure scientific rigor and experimental validity.

\begin{table*}[htbp]
\begin{minipage}{\linewidth}
\centering
\begin{tcolorbox}[colback=blue!0.9!white,colframe=blue!30!black,title=\textbf{Prompt for Merging Parallel Annotation}]
\centering
\footnotesize 
\begin{tabular}{p{\columnwidth} c}

    \VarSty{ {\bf System Message:} } & \\
        \textbf{Input}:& \\
            You are a text integration expert. Here are the parallel annotations of two annotators. Please help me merge their annotations to form a summary annotation.& \\
        \\
        \textbf{Guidelines}:& \\
            $\quad\bullet\quad$ \textbf{Rule 1:} For parts with the same semantic meaning in caption1 and caption2, adopt a merging strategy and avoid repeating the same content.& \\
            $\quad\bullet\quad$ \textbf{Rule 2:} For parts unique to caption1 or caption2, incorporate them into the final description at an appropriate position.& \\
            $\quad\bullet\quad$ \textbf{Rule 3:} For parts where caption1 and caption2 contradict each other, select one caption’s description for the consolidation, and do not include any reference to caption1 or caption2 in the description.& \\
            $\quad\bullet\quad$ \textbf{Rule 4:} During the merging process, avoid redundant mentions of caption1 and caption2. No intermediate reasoning is needed; just provide the final consolidated result.& \\
        \\
        Remember, your output should be a high-quality caption that is concise, informative, and coherent!
        
        \hrulefill & \\
    
    \VarSty{ {\bf User:} } &\\
    
         \#\#\# Caption 1: \textcolor{red}{\texttt{\{first person annotation\}}} \> & \\
         \#\#\# Caption 2: \textcolor{red}{\texttt{\{parallel person annotation\}}} \> & \\
        
        \hrulefill & \\
    
    \VarSty{ {\bf Assistant Generation Prefix:} } & \\
        Here's the merged parallel caption:& \\

\end{tabular}
\end{tcolorbox}

\caption{\label{t10}An Example implementation of the merging parallel annotation function $\sigma_\mathrm{merge}$ via prompting LLMs.}
\end{minipage}
\end{table*}

\begin{table*}[htbp]
\begin{minipage}{\linewidth}
\centering
\begin{tcolorbox}[colback=blue!0.9!white,colframe=blue!30!black,title=\textbf{Prompt for Merging Sequential Annotation}]
\centering
\footnotesize 
\begin{tabular}{p{\columnwidth} c}

    \VarSty{ {\bf System Message:} } & \\
        \textbf{Input}:& \\
            You are a text integration expert. Caption1 is the original annotation result, and Caption2 is the annotator's supplementation and correction.& \\
        \\
        \textbf{Guidelines}:& \\
            $\quad\bullet\quad$ \textbf{Rule 1:} Caption2 will include corrections to Caption1, possibly revising parts of the description in Caption1, as well as supplementing areas where Caption1’s description was insufficient.& \\
            $\quad\bullet\quad$ \textbf{Rule 2:} For parts with the same semantic meaning in Caption1 and Caption2, adopt a merging strategy and avoid repeating the same content.& \\
            $\quad\bullet\quad$ \textbf{Rule 3:} For parts that are missing in Caption1 but present in Caption2, incorporate the relevant parts from Caption2 into Caption1 at an appropriate position.& \\
            $\quad\bullet\quad$ \textbf{Rule 4:} If there is a conflict between the descriptions in Caption1 and Caption2, prioritize the description in Caption2 and replace the corresponding part in Caption1.& \\
        \\
        Remember, your output should be a high-quality caption that is concise, informative, and coherent!
        
        \hrulefill & \\
    
    \VarSty{ {\bf User:} } &\\
    
         \#\#\# Caption 1: \textcolor{red}{\texttt{\{first person annotation\}}} \> & \\
         \#\#\# Caption 2: \textcolor{red}{\texttt{\{sequential person annotation\}}} \> & \\
        
        \hrulefill & \\
    
    \VarSty{ {\bf Assistant Generation Prefix:} } & \\
        Here's the merged sequential caption:& \\

\end{tabular}
\end{tcolorbox}

\caption{\label{t11}An Example implementation of the merging sequential annotation function $\sigma_\mathrm{merge}$ with prompting LLMs.}
\end{minipage}
\end{table*}

\begin{table*}[htbp]
\begin{minipage}{\linewidth}
\centering
\begin{tcolorbox}[colback=blue!0.9!white,colframe=blue!30!black,title=\textbf{Prompt for faster input of text than talk problem acquisition}]
\centering
\footnotesize 
\begin{tabular}{p{\columnwidth} c}
    \VarSty{ {\bf System Message:} } & \\
        \textbf{Input}:& \\
            You are an annotation question-generation assistant. Given a segment of annotation text, please design questions according to the following rules:& \\
        \\
        \textbf{Guidelines}:& \\
            $\quad\bullet\quad$ \textbf{Rule 1:} Generate a total of 5 questions.& \\
            $\quad\bullet\quad$ \textbf{Rule 2:} The five questions should cover the beginning, middle, and later parts of the annotation text.& \\
            $\quad\bullet\quad$ \textbf{Rule 3:} Design the questions in the order of the text and number them sequentially (Q1–Q5).& \\
            $\quad\bullet\quad$ \textbf{Rule 4:} The five questions should progress from general to detailed, starting with broad questions and moving to fine-grained ones.& \\
            $\quad\bullet\quad$ \textbf{Rule 5:} The questions can be about objects or their attributes (\textit{e.g.,} color, quantity, location, shape, size, etc.).& \\
        \\
        \textbf{Example}:& \\
            $\quad\bullet\quad$ \textbf{Question 1:} What kind of image is this describing?  & \\
            $\quad\bullet\quad$ \textbf{Question 2:} What color is the sea surface?& \\
            $\quad\bullet\quad$ \textbf{Question 3:} What's in the top left corner of the picture?& \\
            $\quad\bullet\quad$ \textbf{Question 4:} Where is the parking lot located in the picture?& \\
            $\quad\bullet\quad$ \textbf{Question 5:} Do all houses have swimming pools?& \\
        \\
        
        \hrulefill & \\
    
    \VarSty{ {\bf User:} } &\\
    
         \#\#\# Annotation Text:  \textcolor{red}{\texttt{\{caption\}}} \> & \\
        
        \hrulefill & \\
    
    \VarSty{ {\bf Assistant Generation Prefix:} } & \\
        Here are the generated questions:& \\

\end{tabular}
\end{tcolorbox}

\caption{\label{t14}An Example implementation of the generated questions via prompting LLMs.}

\end{minipage}
\end{table*}

\begin{table*}[htbp]
\begin{minipage}{\linewidth}
\centering
\begin{tcolorbox}[colback=blue!0.9!white,colframe=blue!30!black,title=\textbf{Prompt for Denoising and simplifying annotations}]
\centering
\footnotesize 
\begin{tabular}{p{\columnwidth} c}
    \textbf{Prompt for Denoising and simplifying annotations} & \\
    \VarSty{ {\bf System Message:} } & \\
        \textbf{Input}:& \\
            Please help me improve the following text according to the steps below.& \\
        \\
        \textbf{Guidelines}:& \\
            $\quad\bullet\quad$ \textbf{Rule 1:} Correct obvious types.& \\
            $\quad\bullet\quad$ \textbf{Rule 2:} Remove meaningless connecting words such as "then", "and", "furthermore" and "next".& \\
            $\quad\bullet\quad$ \textbf{Rule 3:} Format the output according to the sample provided.& \\
        \\
        
        \hrulefill & \\
    
    \VarSty{ {\bf User:} } &\\
    
         \#\#\# Annotation Text:  \textcolor{red}{\texttt{\{merged caption\}}} \> & \\
        
        \hrulefill & \\
    
    \VarSty{ {\bf Assistant Generation Prefix:} } & \\
        Here’s the processed caption:& \\

\end{tabular}
\end{tcolorbox}

\caption{\label{t12}Prompt for Denoising and simplifying annotations.}

\end{minipage}
\end{table*}

\begin{table*}[htbp]
\begin{minipage}{\linewidth}
\centering
\begin{tcolorbox}[colback=blue!0.9!white,colframe=blue!30!black,title=\textbf{Prompt for Deriving the minimal Semantic Units}]
\centering
\footnotesize 
\begin{tabular}{p{\columnwidth} c}
    \VarSty{ {\bf Assistant Generation Prefix:} } & \\
        Here’s the processed caption:& \\
            
        \hrulefill & \\
          
    \VarSty{ {\bf System Message:} } & \\
        \textbf{Input}:& \\
            Please help me extract and segment the semantic units according to the following rules and referring to the output example:& \\
        \\
        \textbf{Guidelines}:& \\
            $\quad\bullet\quad$ \textbf{Unit Definition:} Semantic unit = object name + associated attributes; A single sentence may contain multiple independent units; Each unit must contain only one object name.& \\
            $\quad\bullet\quad$ \textbf{Attribute Specifications:} Valid attributes: absolute\_location (position in the overall image), relative\_location(Position relative to other objects), colour, amount (Explicitly extract indefinite articles "a"/"an" as standalone attributes. Include numerical values \textit{e.g.,} "two", "three" and quantifiers (\textit{e.g.,} "some", "several")), size, shape, material, object description, other(All other unclassified attributes are 'other', If there are multiple, please separate them with commas),Omit any attributes that do not exist; Prohibit attribute overlap or duplication; Pronoun-based locations (\textit{e.g.,} "this", "that") must be replaced with specific referenced objects.& \\
            $\quad\bullet\quad$ \textbf{Extraction Principles:} Prioritize extracting the "name" field separately; Create independent units for multiple objects sharing attributes; Absolute and relative locations cannot coexist in the same unit; Omit unspecified/ambiguous attributes.& \\
            $\quad\bullet\quad$ \textbf{Output Requirements:} Present only final results without reasoning processes.& \\
            
        \\
        \textbf{Example}:& \\
            $\quad\bullet\quad$ \textbf{Input Example 1:} 
            The sea surface appears green, with a patch of green seaweed visible under the bridge in the upper right area.  & \\
            $\quad\bullet\quad$ \textbf{Output Example 1:} 
    \begin{verbatim}
    [
        {
            "name": "sea surface",
            "attributes": {
                "colour": "green",
                "other": ["appears"]
            }
        },
        {
            "name": "seaweed",
            "attributes": {
                "amount": "a patch of",
                "colour": "green",
                "relative_location": "under the bridge in the upper right area",
                "other": ["visible"]
            }
        }
    ]
    …
    \end{verbatim}
    & \\
        
        \hrulefill & \\
    
    \VarSty{ {\bf User:} } &\\
    
         \#\#\# Caption:  \textcolor{red}{\texttt{\{processed caption\}}} \> & \\
        
        \hrulefill & \\
    
    \VarSty{ {\bf Assistant Generation Prefix:} } & \\
        Here are the Semantic Units:& \\
        
\end{tabular}
\end{tcolorbox}

\caption{\label{t13}Prompt for Deriving the minimal Semantic Units.}

\end{minipage}
\end{table*}

\begin{figure*}[htbp]
    \centering
    \includegraphics[width=1\linewidth]{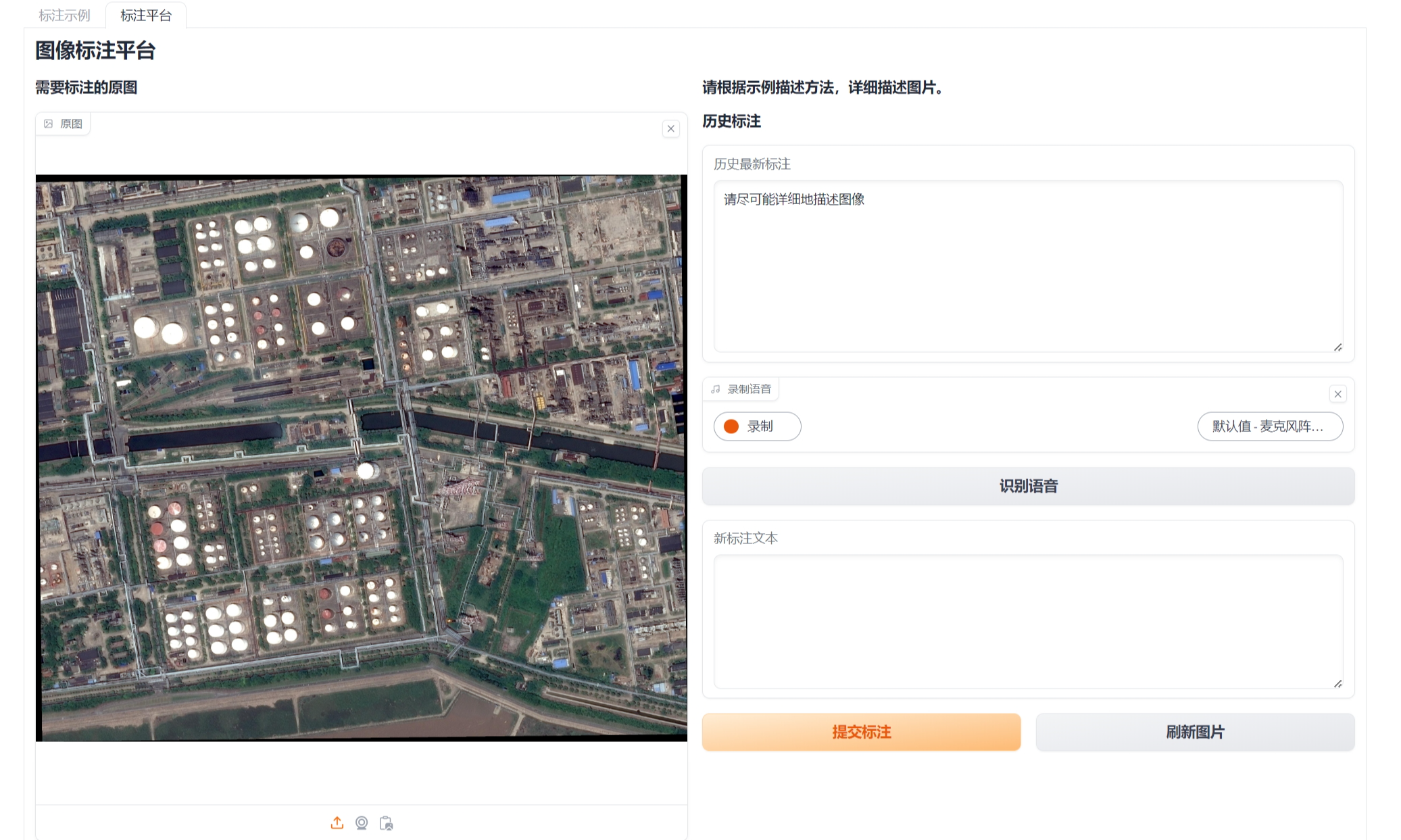}
    \caption{The first-person annotation interface.}
    \label{fig:interface1}
\end{figure*}

\begin{figure*}[htbp]
    \centering
    \includegraphics[width=1\linewidth]{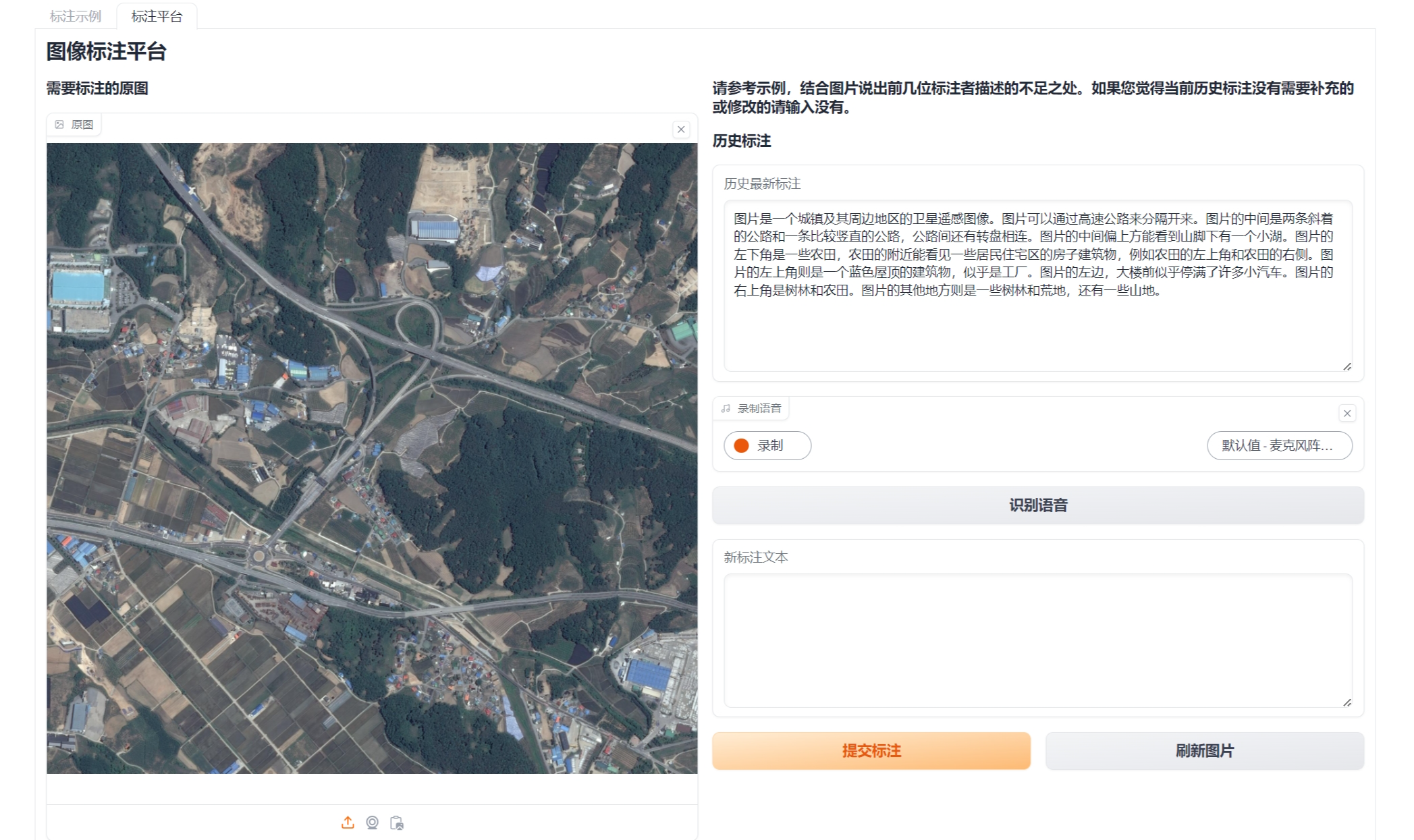}
    \caption{The Suquential Subsequent Annotation Interface.}
    \label{fig:interface2}
\end{figure*}

\subsection{Derive the Semantic Units}
\label{Derive the Semantic Units}

For the intrinsic evaluation of image caption quality, we employ a text-based LLM to decompose image captions into their semantic units and assess them according to the number of semantic units identified. Following the procedure outlined in Table \ref{t12} and Table \ref{t13}, we first perform necessary simplification and standardization of the combined text from Section~\ref{Merge Annotation}, then systematically split it to extract all the semantic units contained within the captions.

\section{Annotation Interface}
\begin{table*}[t]
\begin{minipage}{\linewidth}
\centering
\begin{tcolorbox}[colback=blue!0.9!white,colframe=blue!30!black,title=\textbf{Guidelines for Image Annotation}]
\centering
\footnotesize 
\begin{tabular}{p{\columnwidth} c}
    \textbf{First-Person:}& \\
    \VarSty{ {\bf System Message:} } & \\
        \textbf{Input}:& \\
            You will be provided with an image. Your task is to generate a detailed and informative caption for the image, adhering to the following guidelines: & \\
        \textbf{Guidelines}:& \\
            $\quad\bullet\quad$ \textbf{Rule 1:} The caption should be as comprehensive as possible. Identify and describe all discernible entities in the image along with their attributes. Attributes may include (but are not limited to): Absolute position (via image orientation), Relative position (in reference to other objects), Color, quantity, size, shape, material, etc. Avoid describing entities that cannot be clearly identified.& \\
            $\quad\bullet\quad$ \textbf{Rule 2:} Structure the caption in the following order: First: Begin with a global description of the entire image; Second, Provide a description of the objects located at the center of the image. Last, Describe the entire image systematically, starting from the upper-left corner and proceeding in a structured manner with the spatial relationships between objects.& \\
            $\quad\bullet\quad$ \textbf{Rule 3:} If there are more than 10 instances of a particular object type, leverage approximate quantifiers (\textit{e.g.,} many, some, a row, a column, a cluster, etc.) instead of exact counts.& \\
            $\quad\bullet\quad$ \textbf{Rule 4:} Ensure the caption is concise but information-rich. Each sentence should contain meaningful and non-redundant information. Avoid vague, repetitive, or empty expressions.& \\
            \hrulefill & \\
    \\
    \textbf{Subsequent-Person:}& \\
    \VarSty{ {\bf System Message:} } & \\
        \textbf{Input}:& \\
            You will be provided with an image and its corresponding caption. Your task is to review and revise the caption to ensure it accurately and comprehensively reflects the content of the image, following the rules below:  & \\
        \textbf{Guidelines}:& \\
            $\quad\bullet\quad$ \textbf{Rule 1:} Examine whether the caption includes all identifiable entities present in the image, along with their corresponding attributes. Attributes may include (but are not limited to): Absolute position (according to image orientation) Relative position (with respect to other objects) Color, quantity, size, shape, material, etc. If any entity is missing, add it along with its attributes. If any attribute of a described entity is missing, supplement it accordingly.& \\
            $\quad\bullet\quad$ \textbf{Rule 2:} If the caption contains any inaccuracies (\textit{e.g.,} incorrect quantity, color, or other attributes of an entity), make the necessary corrections. & \\
            $\quad\bullet\quad$ \textbf{Rule 3:} Output only the revised caption. Do not include or refer to the original caption in your response.& \\    
            \\
\end{tabular}
\end{tcolorbox}

\caption{\label{t1_2}Guidelines for Image Annotation.}

\end{minipage}
\end{table*}

\subsection{Annotation Interface 1: The first-person Annotation Interface of the CoTalk framework}
The first annotator is tasked with generating detailed image captions that comprehensively describe the visual content. These captions are provided via voice input, transcribed into text, and then standardized using a large language model for storage. Each caption should aim to cover all identifiable entities in the image along with their attributes, including—but not limited to—absolute and relative positions, color, quantity, size, shape, and material. Details are shown in Table \ref{t1_2} (up).

\subsection{Annotation Interface 2: The Suquential Subsequent Annotation Interface of CoTalk Framework}

In CoTalk, subsequent annotators review the image and previously merged captions, generated by a large model, to identify omissions and correct errors. They contribute additional information through voice input, guided by a comprehensive understanding of both the image and the existing annotations.Details are shown in Table \ref{t1_2} (down).

\begin{figure*}[t]
    \centering
    \includegraphics[width=1\linewidth]{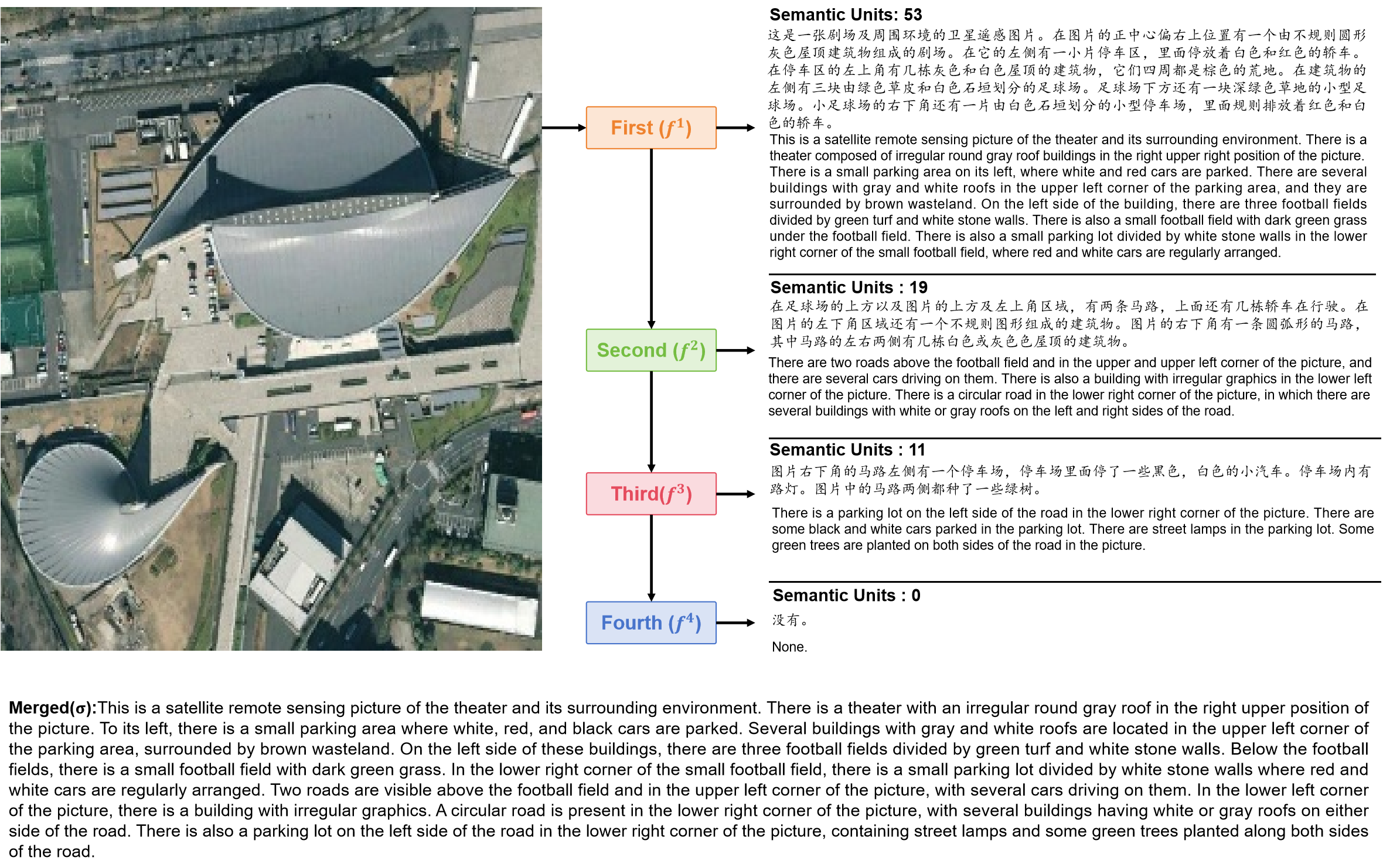}
    \caption{Example of image annotation in the area around the theatre.}
    \label{fig:Example1}
\end{figure*}
\section{Detailed Experimental Process}

\subsection{Extrinsic Metric}
We evaluate the practicality of each annotation method using extrinsic metrics that indirectly reflect annotation quality, employing a retrieval task for this purpose. Specifically, we fine-tune the Long-CLIP model on datasets annotated by CoTalk and a parallel method. Long-CLIP is selected for its extended input capacity of 248 tokens, nearly four times that of the original CLIP (77 tokens), making it well-suited for capturing detailed annotations.

We fine-tune the Long-CLIP-L model using three V100 GPUs, with a batch size of 16, a learning rate of 1e-6, and 12 epochs. Each epoch takes approximately 18 seconds. To ensure robust results, we run five trials with different random seeds per dataset and report the average performance.

After training, we evaluate the model on a remote sensing retrieval dataset. As shown in Table \ref{complete_t1}, CoTalk achieves 65.94\% across six tasks and three datasets, outperforming the parallel method (65.01\%) and the zero-shot baseline (62.54\%).
\label{app:Extrinsic Metric}

\subsection{Consistency between Extrinsic and Intrinsic Metrics}
To verify the consistency between intrinsic indicators (\textit{i.e.,} the number of semantic units) and extrinsic indicators (\textit{i.e.,} downstream task performance), we reduce the number of semantic units in the CoTalk-annotated dataset of 429 images, at fixed ratios. Specifically, for each image, we randomly remove 20\%, 30\%, 50\%, 70\%, or 80\% of its Semantic Units. For example, if an image originally has 10 Semantic Units and 20\% are removed, 2 are randomly deleted, and the remaining 8 are merged into a single text input for subsequent fine-tuning of the Long-CLIP model.

We fine-tune Long-CLIP-L using three V100 GPUs with a batch size of 16, a learning rate of 1e-6, and 12 epochs. To ensure stability, we repeat each experiment using five different random seeds per dataset and report the average results.

We evaluate performance on the RSICD, RSITMD, and UCM-Captions retrieval datasets. As shown in Table~\ref{complete_t2}, retrieval performance declines as fewer Semantic Units are retained, confirming the alignment between intrinsic and extrinsic indicators. This supports the use of Semantic Units as an effective measure of annotation quality and practical utility. Notably, when more than 50\% of Semantic Units are removed, the performance drop becomes more pronounced, indicating the importance of detailed captions for effective vision-language alignment.
\label{app: Consistency}

\section{CoTalk Examples}

\begin{figure*}[h]
    \centering
    \includegraphics[width=1\linewidth]{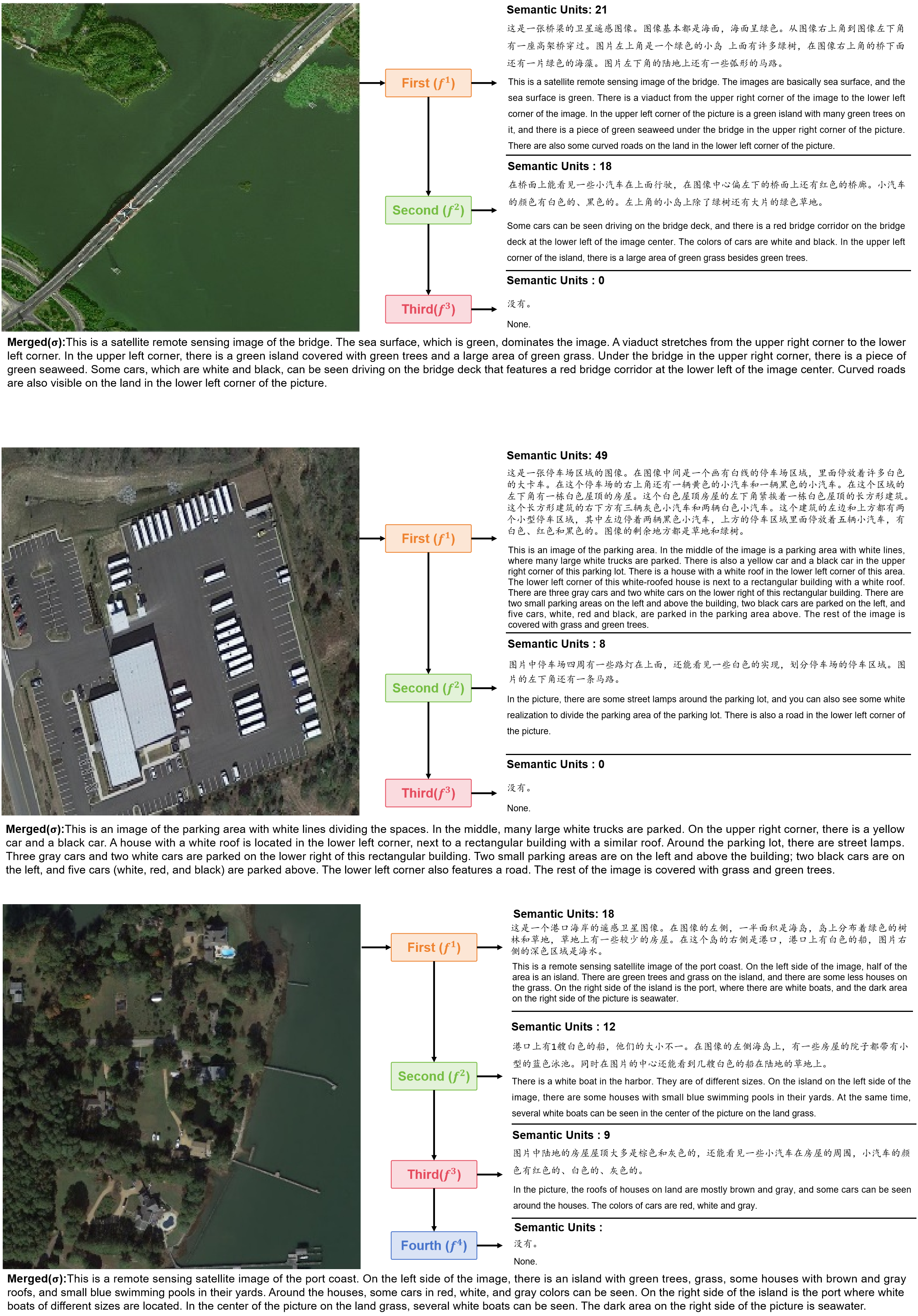}
    \caption{Examples of image annotation of bridge periphery (top), parking lot periphery (middle) and port coast (bottom).}
    \label{fig:Example2}
\end{figure*}

This section presents representative image samples manually annotated to demonstrate the labeling process. As illustrated in Figure \ref{fig:Example1} and Figure \ref{fig:Example2}, the examples span typical scenes: (1) theaters and surrounding urban areas, (2) bridges and adjacent suburban waters, (3) parking lots and their environments, (4) port piers with coastal landscapes. These samples reflect both the annotation quality and the integration of large model predictions. The resulting text data accurately reflect the spatial structures and semantic content of each scene, underscoring the reliability of our annotations.

\end{document}